\documentclass[ba,preprint]{imsart}
\pubyear{2024}
\volume{TBA}
\issue{TBA}
\firstpage{1}
\lastpage{1}

\usepackage{amsthm}
\usepackage{amsmath}
\usepackage{natbib}
\usepackage[colorlinks,citecolor=blue,urlcolor=blue,filecolor=blue,backref=page]{hyperref}
\usepackage{graphicx}

\startlocaldefs
\numberwithin{equation}{section}
\theoremstyle{plain}

\endlocaldefs

\usepackage{amsmath,amsthm}
\usepackage{amssymb}
\usepackage{bbm}
\usepackage{amsfonts}
\usepackage{graphicx}
\usepackage{subfigure}
\graphicspath{{graphics/}}
\usepackage[ruled,vlined,linesnumbered]{algorithm2e}
\usepackage{algorithmic}
\usepackage{tabularx,booktabs}
\usepackage{tikz}
\usepackage{xcolor}
\definecolor{darkblue}{rgb}{0.0,0.5,0.5}
\usepackage{booktabs,caption}
\usepackage{multirow}
\usepackage{lscape}
\usepackage{epstopdf}
\usepackage{lineno}
\usepackage{subfigure}

\newcommand{\given}{\;\middle|\;}
\newcommand{\R}{\mathbb{R}}
\newcommand{\bd}[1]{\boldsymbol{#1}}

\begin{document}

\begin{frontmatter}
\title{Scalable Spatiotemporally Varying Coefficient Modeling with Bayesian Kernelized Tensor Regression}
\runtitle{Scalable STVC Modeling with BKTR}

\begin{aug}
\author{\fnms{Mengying} \snm{Lei}\thanksref{addr1}\ead[label=e1]{mengying.lei@mail.mcgill.ca,}\orcid{0000-0001-7343-3323}},
\author{\fnms{Aur\'elie} \snm{Labbe}\thanksref{addr2}\ead[label=e3]{aurelie.labbe@hec.ca.}\orcid{0009-0008-2182-0637}}
\and
\author{\fnms{Lijun} \snm{Sun}\thanksref{addr1,t1}%
\ead[label=e2]{lijun.sun@mcgill.ca.}\orcid{0000-0001-9488-0712}%
}

\runauthor{M. Lei, A. Labbe, and L. Sun}

\address[addr1]{Department of Civil Engineering, McGill University, Montreal, Quebec, H3A 0C3, Canada,
    \printead{e1} 
    \printead*{e2}
}

\address[addr2]{Department of Decision Sciences, HEC Montr\'eal, Montreal, Quebec, H3T 2A7, Canada,
    \printead{e3}
}

\thankstext{t1}{Corresponding author}

\end{aug}

\begin{abstract}
As a regression technique in spatial statistics, the spatiotemporally varying coefficient model (STVC) is an important tool for discovering nonstationary and interpretable response-covariate associations over both space and time. However, it is difficult to apply STVC for large-scale spatiotemporal analyses due to its high computational cost. To address this challenge, we summarize the spatiotemporally varying coefficients using a third-order tensor structure and propose to reformulate the spatiotemporally varying coefficient model as a special low-rank tensor regression problem. The low-rank decomposition can effectively model the global patterns of large data sets with a substantially reduced number of parameters. To further incorporate the local spatiotemporal dependencies, we use Gaussian process (GP) priors on the spatial and temporal factor matrices. We refer to the overall framework as Bayesian Kernelized Tensor Regression (BKTR), and kernelized tensor factorization can be considered a new and scalable approach to modeling multivariate spatiotemporal processes with a low-rank covariance structure. For model inference, we develop an efficient Markov chain Monte Carlo (MCMC) algorithm, which uses Gibbs sampling to update factor matrices and slice sampling to update kernel hyperparameters. We conduct extensive experiments on both synthetic and real-world data sets, and our results confirm the superior performance and efficiency of BKTR for model estimation and parameter inference.
\end{abstract}


\end{frontmatter}

\section{Introduction}
\label{sec:intro}

Local spatial regression aims to characterize the nonstationary and heterogeneous associations between the response variable and the corresponding covariates observed in a spatial domain \citep{banerjee2014hierarchical,cressie2015statistics}. This is achieved by assuming that the regression coefficients vary locally over space. Local spatial regression offers enhanced interpretability of complex relationships and has become an important technique in many fields, such as geography, ecology, economics, environment, public health and climate science, to name but a few. In general, a local spatial regression model for a scalar response $y$ can be written as:
\begin{equation}\label{eq:svc}
    y(\boldsymbol{s}) = \bd{x}(\boldsymbol{s})^{\top} \bd{\beta}(\bd{s})+ \epsilon(\bd{s}),
\end{equation}
where $\bd{s}$ is the index (e.g., longitude and latitude) for a spatial location, $\bd{x}(\bd{s})\in\mathbb{R}^P$ and $\bd{\beta}(\bd{s})\in\R^{P}$ are the covariate vector and the regression coefficients at location $\bd{s}$, respectively, and $\epsilon(\bd{s})\sim \text{i.i.d.}\,\mathcal{N}(0,\tau^{-1})$ is a white noise process with precision $\tau$.

There are two common methods for local spatial regression analysis---the Bayesian spatially varying coefficient model (SVC) \citep{gelfand2003spatial} and the geographically weighted regression (GWR) \citep{fotheringham2003geographically}. SVC is a Bayesian hierarchical model in which the regression coefficients are modeled using Gaussian processes (GP) with a kernel function to be learned \citep{rasmussen2006}. For a collection of $M$ observed locations, the original SVC
 developed by \citet{gelfand2003spatial} imposes a prior such that $\text{vec}(\bd{\beta}_{\text{mat}}^{\top})\sim \mathcal{N} (\bd{1}_{M\times 1}\otimes\bd{\mu}_{\beta}, \bd{K}_s\otimes \bd{\Lambda}^{-1})$, where $\bd{\beta}_{\text{mat}}$ is a $M\times P$ matrix of all coefficients, $\text{vec}(\bd{X})$ denotes vectorization by stacking all columns in $\bd{X}$ as a vector, $\bd{\mu}_{\beta}$ represents the overall regression coefficient vector used to construct the mean,
 $\bd{K}_s$ is a $M\times M$ spatial correlation matrix, $\bd{\Lambda}$ is a $P\times P$ precision matrix for covariates, and $\otimes$ denotes the Kronecker product.
 In this paper, for simplicity, we use a zero-mean GP to specify $\bd{\beta}$, and the global effect of the covariates can be learned (or removed) through a linear regression term as in \citet{gelfand2003spatial}. In addition, setting $\bd{K}_s$ as a correlation matrix simplifies the covariance specification since the variance can be captured by scaling $\bd{\Lambda}^{-1}$. This formulation is equivalent to having a matrix normal distribution $\bd{\beta}_{\text{mat}}\sim \mathcal{MN}_{M\times P}\left(\bd{0},\bd{K}_s,\bd{\Lambda}^{-1}\right)$. GWR was developed independently using a local weighted regression approach, in which a bandwidth parameter is used to calculate the weights (based on a weight function) for different observations, with closer observations carrying larger weights. In practice, the bandwidth parameter is either prespecified based on domain knowledge or tuned through cross-validation. However, it has been shown in the literature that the estimation results of GWR are highly sensitive to the selection of the bandwidth parameter \citep[e.g.,][]{finley2011comparing}. Compared with GWR, the Bayesian hierarchical framework of SVC provides more robust results and allows us to learn the hyperparameters of the spatial kernel $\bd{K}_s$, e.g., the length-scale, which is critical to understanding the underlying characteristics of the spatial processes. In addition, by using Markov chain Monte Carlo (MCMC), we can not only explore the posterior distribution of the kernel hyperparameters and regression coefficients, but also perform out-of-sample prediction with uncertainty quantification.

The formulation in Eq.~\eqref{eq:svc} can be easily extended to local spatiotemporal regression to further characterize the temporal variation of the coefficients. For a response matrix  $\boldsymbol{Y}\in\mathbb{R}^{M\times N}$ observed from a set of locations $S=\{\bd{s}_1,\ldots,\bd{s}_M\}$ over a set of time points $T=\{t_1,\ldots,t_N\}$, the local spatiotemporal regression model defined on the Cartesian product $S \times T = \left\{\left(\bd{s}_m, t_n\right) : m =
1,\ldots, M, \  n = 1,\ldots, N\right\}$ can be formulated as:
\begin{equation} \label{Eq:STVC-point}
    y\left(\bd{s}_m,t_n\right)=\boldsymbol{x}\left(\bd{s}_m,t_n\right)^{\top}\boldsymbol{\beta}\left(\bd{s}_m,t_n\right)+\epsilon\left(\bd{s}_m,t_n\right),
\end{equation}
where we use $m=1,\ldots,M$ and $n=1,\ldots,N$ to index rows (i.e., location) and columns (i.e., time point), respectively, $y\left(\bd{s}_m,t_n\right)$ is the $(m,n)$th element in $\bd{Y}$, and $\boldsymbol{x}\left(\bd{s}_m,t_n\right)$  and $\bd{\beta}\left(\bd{s}_m,t_n\right)$ are the covariate vector and coefficient vector at location $\bd{s}_{m}$ and time $t_{n}$, respectively. Based on this formulation, \citet{huang2010geographically} extended GWR to geographically and temporally weighted regression (GTWR) by introducing more parameters to quantify spatiotemporal weights in the locally weighted regression. For SVC, \citet{gelfand2003spatial} suggested using a separable kernel structure to build a spatiotemporally varying coefficient model (STVC), which assumes that $\left[\boldsymbol{\beta}\left(\bd{s}_1,t_1\right);\ldots;\bd{\beta}\left(\bd{s}_M,t_1\right);\bd{\beta}\left(\bd{s}_1,t_2\right);\ldots;\boldsymbol{\beta}\left(\bd{s}_M,t_N\right) \right]\sim \mathcal{N}(\bd{0}, \bd{K}_t\otimes \bd{K}_s \otimes \bd{\Lambda}^{-1})$, where $\bd{K}_t$ is a $N\times N$ kernel matrix defining the correlation structure for the $N$ time points. Note that with this GP formulation, it is not necessary for the $N$ time points to be {\color{blue}equally spaced}. If we parameterize the regression coefficients in Eq.~\eqref{Eq:STVC-point} as a third-order tensor $\bd{\mathcal{B}}\in \R^{M\times N\times P}$ with mode-3 fiber $\bd{\mathcal{B}}(m,n,:)=\boldsymbol{\beta}\left(\bd{s}_m,t_n\right)$, the above specification is equivalent to having a tensor normal distribution $\bd{\cal{B}}\sim \mathcal{TN}_{M\times N\times P}\big(\bd{0},\bd{K}_s,\bd{K}_t,$ $\bd{\Lambda}^{-1}\big)$. However, despite the elegant separable kernel-based formulation in STVC, the model is rarely used in real-world practice mainly due to the high computational cost. For example, for a fully observed matrix $\bd{Y}$ with corresponding spatiotemporal covariates, updating the coefficients $\bd{\beta}$ in each MCMC iteration requires time complexity of $\mathcal{O}\left(M^3N^3P^3\right)$. Updating the kernel hyperparameters can be achieved by integrating out $\bd{\beta}$, but it still requires $\mathcal{O}\left(M^3N^3\right)$ in time.

In this paper, we provide an alternative estimation strategy---Bayesian Kernelized Tensor Regression (BKTR)---to perform Bayesian spatiotemporal regression analysis on large-scale data sets. Inspired by the idea of low-rank regression and tensor regression \cite[see e.g.,][]{izenman1975reduced,cressie2008fixed,banerjee2008gaussian,zhou2013tensor,bahadori2014fast,guhaniyogi2017bayesian}, we use low-rank tensor factorization to encode the dependencies among the three dimensions in $\bd{\mathcal{B}}$ with only a few latent factors. To further incorporate local spatial and temporal dependencies, we use GP priors on the spatial and temporal factor vectors following \citet{gamerman2008spatial} and \citet{luttinen2009variational}, thus translating the default tensor factorization into a kernelized factorization model. With a specified tensor rank $R$, the time complexity becomes $\mathcal{O}\left(R^3\left(M^3+N^3+P^3\right)\right)$, which is substantially reduced compared with the default STVC formulation. In addition to the spatial and temporal framework, we also consider the case where a proportion of the response matrix $\boldsymbol{Y}$ can be unobserved or corrupted, given observed values of the covariates $\boldsymbol{X}$. Such a scenario is very common in many real-world applications, such as traffic state data collected from emerging crowdsourcing and moving sensing systems (e.g., Google Waze) for example, where observations are inherently sparse in space and time. We show that the underlying Bayesian tensor decomposition structure allows us to effectively estimate both the model coefficients and the unobserved outcomes even when the missing rate of $\boldsymbol{Y}$ is high. We conduct numerical experiments on both synthetic and real-world data sets, and our results confirm the promising performance of BKTR.

\section{Related Work}
\label{sec:RelatedWork}
The key computational challenge in SVC/STVC is how to efficiently and effectively learn a multivariate spatial/spatiotemporal process (i.e., $\bd{\beta}_{\text{mat}}$ in Eq.~\eqref{eq:svc} and the third-order spatiotemporal tensor $\bd{\cal{B}}$ in Eq.~\eqref{Eq:STVC-point}). For a general multivariate spatial process that is fully observed on the Cartesian product with white noise, a popular approach is to use separable covariance specification on which one can leverage the property of Kronecker products to substantially reduce the computational cost \citep{saatcci2012scalable, wilson2014fast}. However, for SVC, we cannot benefit directly from the Kronecker property since the data $\bd{y}$ is obtained through a linear transformation of $\bd{\beta}_{\text{mat}}$. In this case, computing the inverse of an $MP\times MP$ matrix becomes inevitable when sampling these spatially varying coefficients. Existing frameworks for SVC essentially adopt a two-step approach \citep{gelfand2003spatial,finley2020bayesian}: (1) update only kernel hyperparameters and $\bd{\mu}$ by marginalizing $\bd{\beta}$ with cost $\mathcal{O}\left(M^3\right)$; and (2) after burn-in, use composition sampling on the obtained MCMC samples to generate samples for $\bd{\beta}$ with cost $\mathcal{O}\left(M^3P^3\right)$; see \citet{finley2020bayesian} for a detailed implementation of Bayesian SVC. For STVC, the corresponding costs in the two steps are $\mathcal{O}\left(M^3N^3\right)$ and $\mathcal{O}\left(M^3N^3 P^3\right)$, respectively. The high computational cost in step (2) is the primary issue that limits the application of SVC/STVC in practice.

Our work follows a different approach. Instead of modeling $\bd{\beta}$ directly using a GP, we parameterize the third-order tensor $\bd{\cal{B}}$ for STVC using a low-rank tensor decomposition \citep{kolda2009tensor}. The idea is inspired by recent studies on low-rank tensor regression/learning \cite[see e.g.,][]{gamerman2008spatial,zhou2013tensor,bahadori2014fast,rabusseau2016low,yu2016learning,guhaniyogi2017bayesian,yu2018tensor}. The low-rank assumption not only preserves the global patterns and higher-order dependencies in the variable, but also greatly reduces the number of parameters. In fact, without considering spatiotemporal indices, we can formulate Eq.~\eqref{Eq:STVC-point} as a scalar-tensor regression problem \citep{guhaniyogi2017bayesian} by reconstructing each $\bd{x}(\bd{s}_m,t_n)$ as a sparse covariate tensor of the same size as $\bd{\cal{B}}$. However, for spatiotemporal data, the low-rank assumption alone cannot fully characterize the strong local spatial and temporal consistency. To better encode local spatial and temporal consistency, existing studies in tensor regression have introduced graph Laplacian regularization in defining the loss function \cite[e.g.,][]{bahadori2014fast,rao2015collaborative} in an optimization framework. Nevertheless, this approach also introduces more parameters (e.g., those used to define the distance/similarity function and weights in the loss function) and without a Bayesian hierarchical specification it has limited power in modeling complex spatial and temporal processes. The most relevant work is a Gaussian process factor analysis model \citep{luttinen2009variational} developed for a completely different problem---completing a spatiotemporal matrix observed from $M$ locations over $N$ time points, in which different GP priors are assumed on the spatial and temporal factors, and the whole model is learned through variational Bayesian inference. Similarly, \citet{gamerman2008spatial} developed a spatial dynamic factor model in which spatial factors are assumed to have GP priors and temporal factors follow a dynamic linear model. \citet{lei2022bayesian} presents an MCMC scheme for this model, in which slice sampling is used to update kernel hyperparameters and Gibbs sampling is used to update factor matrices. We follow a similar idea as in \citet{luttinen2009variational} and \citet{lei2022bayesian} to parameterize the coefficients $\bd{\beta}$ and develop MCMC algorithms for model inference. In a recent work, \citet{zhang2022spatial} also proposed to use a Bayesian Linear Model of Coregionalization (LMC) factor model to model high-dimensional multivariate spatial processes involving both $M$ and $P$. To solve the large $M$ issue, the Nearest Neighbor Gaussian Process (NNGP) \citep{datta2016hierarchical,finley2019efficient} is used to model spatial factors. In the literature, there also exist other parameterization methods to model spatial processes/coefficients with multidimensional structures. For instance, \citet{martinez2017towards} used Kronecker decomposition to model a large coefficient matrix for tensor-variate data; \citet{guhaniyogi2023bayesian} proposed to model spatially varying coefficients using basis-function models (with pre-defined basis functions and random basis coefficients) and used sketching to reduce data dimensionality to achieve scalable inference for large $M$. BKTR can also be considered to use tensor factorization as a special basis function method to model $\boldsymbol{\mathcal{B}}$, where the basis functions are also random variables (see \citet{lei2023bayesian} and \citet{cressie2022basis}).

\section{Bayesian Kernelized Tensor Regression}
\label{sec:meth}
\subsection{Preliminaries}
\label{sec:pre}
\paragraph{Notations}
Throughout this paper, we use lowercase letters to denote scalars, e.g., $x$, boldface lowercase letters to denote vectors, e.g., $\boldsymbol{x}\in\mathbb{R}^{M}$, and boldface uppercase letters to denote matrices, e.g., $\boldsymbol{X}\in\mathbb{R}^{M\times N}$. The $\ell_{2}$-norm of $\boldsymbol{x}$ is defined as $\|\boldsymbol{x}\|_{2}=\sqrt{\sum_{m}x_{m}^{2}}$. For a vector $\bd{x}$, we denote its $m$th entry by $\bd{x}(m)$. For a matrix $\boldsymbol{X}\in\mathbb{R}^{M\times N}$, we denote its $(m,n)$th entry by $x_{m,n}$ or $\bd{X}(m,n)$. We use $\bd{I}_{N}$ to denote an identity matrix of size $N\times N$. Given two matrices $\boldsymbol{A}\in\mathbb{R}^{M\times N}$ and $\boldsymbol{B}\in\mathbb{R}^{P\times Q}$, the Kronecker product is defined as $\boldsymbol{A}\otimes\boldsymbol{B}=
\footnotesize{\begin{bmatrix}
  a_{1,1}\boldsymbol{B} & \cdots & a_{1,N}\boldsymbol{B} \\
  \vdots & \ddots & \vdots \\
  a_{M,1}\boldsymbol{B} & \cdots & a_{M,N}\boldsymbol{B}
\end{bmatrix}}\in\mathbb{R}^{MP\times NQ}$. If $\boldsymbol{A}=[\boldsymbol{a}_{1},\ldots,\boldsymbol{a}_{N}]$ and $\boldsymbol{B}=[\boldsymbol{b}_{1},\ldots,\boldsymbol{b}_{Q}]$ have the same number of columns, i.e., $N=Q$, then the Khatri-Rao product is defined as the column-wise Kronecker product $\boldsymbol{A}\odot\boldsymbol{B}=[\boldsymbol{a}_{1}\otimes\boldsymbol{b}_{1},\ldots,\boldsymbol{a}_{N}\otimes\boldsymbol{b}_{N}]\in\mathbb{R}^{MP\times N}$. The vectorization $\operatorname{vec}(\bd{X})$ stacks all column vectors in $\bd{X}$ as a single vector. Following the tensor notation in \citet{kolda2009tensor}, we denote a third-order tensor by $\boldsymbol{\mathcal{X}}\in\mathbb{R}^{M\times N\times P}$ and its mode-$k$ $(k=\text{1},\text{2},\text{3})$ unfolding by $\boldsymbol{X}_{(k)}$, which maps a tensor into a matrix. The mode-3 fibers and frontal slices of $\boldsymbol{\mathcal{X}}$ are denoted by $\boldsymbol{\mathcal{X}}(m,n,:)\in\mathbb{R}^{p}$ and $\boldsymbol{\mathcal{X}}(:,:,p)\in\mathbb{R}^{M\times N}$, respectively. Lastly, we use $\operatorname{ones}(M,N)$ and $\boldsymbol{1}_{M}\in\mathbb{R}^{M}$ to represent a $M\times N$ matrix and a length $M$ column vector of ones, respectively.

\paragraph{Tensor CP decomposition}
For a third-order tensor $\boldsymbol{\mathcal{A}}\in\mathbb{R}^{M\times N\times P}$, the CANDECOMP/PARAFAC (CP) decomposition factorizes $\boldsymbol{\mathcal{A}}$ into a sum of rank-one tensors \citep{kolda2009tensor}:
\begin{equation} \label{Eq:CP decomposition}
    \boldsymbol{\mathcal{A}}=\sum_{r=1}^{R}\boldsymbol{u}_{r}\circ\boldsymbol{v}_{r}\circ\boldsymbol{w}_{r},
\end{equation}
where $R$ is the CP rank, $\circ$ represents the outer product, $\boldsymbol{u}_{r}\in\mathbb{R}^{M}$, $\boldsymbol{v}_{r}\in\mathbb{R}^{N}$, and $\boldsymbol{w}_{r}\in\mathbb{R}^{P}$ for $r=1,\ldots,R$. The factor matrices that combine the vectors from the rank-one components are denoted by $\boldsymbol{U}=[\boldsymbol{u}_{1},\ldots,\boldsymbol{u}_{R}]\in\mathbb{R}^{M\times R}$, $\boldsymbol{V}\in\mathbb{R}^{N\times R}$, and $\boldsymbol{W}\in\mathbb{R}^{P\times R}$, respectively. We can write Eq.~\eqref{Eq:CP decomposition} in the following matricized form:
\begin{equation} \label{Eq:unfolding-CP}
    \begin{split}
        \boldsymbol{A}_{(1)}=\boldsymbol{U}\left(\boldsymbol{W}\odot\boldsymbol{V}\right)^{\top},~
        \boldsymbol{A}_{(2)}=\boldsymbol{V}\left(\boldsymbol{W}\odot\boldsymbol{U}\right)^{\top},~
        \boldsymbol{A}_{(3)}=\boldsymbol{W}\left(\boldsymbol{V}\odot\boldsymbol{U}\right)^{\top},
    \end{split}
\end{equation}
where $\odot$ is the Khatri-Rao product. Eq.~\eqref{Eq:unfolding-CP} relates the mode-$k$ unfolding of a tensor to its polyadic decomposition.

\subsection{Model specification}
Let $\boldsymbol{\mathcal{X}}$ be an $M\times N\times P$ tensor, of which the $(m,n)$th mode-3 fiber is the covariate vector at location $\bd{s}_{m}$ and time $t_n$, i.e., $\boldsymbol{\mathcal{X}}{(m,n,:)}=\boldsymbol{x}\left(\bd{s}_m,t_n\right)$. For example, in the application of spatiotemporal modeling on bike-sharing demand that we illustrate later (see Section~\ref{sec:application}), the response matrix $\boldsymbol{Y}\in\mathbb{R}^{M\times N}$ is a matrix of daily departure trips for $M$ bike stations over $N$ days, and the tensor variable $\boldsymbol{\mathcal{X}}\in\mathbb{R}^{M\times N\times P}$ represents a set of $P$ spatiotemporal covariates for the corresponding locations and time. Using $\boldsymbol{y}\in\mathbb{R}^{MN}$ to denote $\operatorname{vec}(\boldsymbol{Y})$, Eq. (\ref{Eq:STVC-point}) can be formulated as:
\begin{equation} \label{Eq:STVC-vector}\boldsymbol{y}=\left(\boldsymbol{I}_{MN}\odot\boldsymbol{X}_{(3)}\right)^{\top}\operatorname{vec}\left(\boldsymbol{B}_{(3)}\right)+\boldsymbol{\epsilon},
\end{equation}
where $\boldsymbol{X}_{(3)}$ and $\boldsymbol{B}_{(3)}$ are the unfoldings of $\boldsymbol{\mathcal{X}}$ and $\boldsymbol{\mathcal{B}}$, respectively, the Khatri-Rao product $\left(\boldsymbol{I}_{MN}\odot\boldsymbol{X}_{(3)}\right)^{\top}$ is a $MN\times MNP$ block diagonal matrix, and $\boldsymbol{\epsilon}\sim\mathcal{N}(\boldsymbol{0},\tau^{-1}\boldsymbol{I}_{MN})$. Assuming that
\begin{equation} \label{Eq:CPdecompB}
\boldsymbol{\mathcal{B}}=\sum_{r=1}^{R}\boldsymbol{u}_{r}\circ\boldsymbol{v}_{r}\circ\boldsymbol{w}_{r}
\end{equation}
admits a CP decomposition with rank $R\ll\min\{M,N\}$, we can rewrite Eq. \eqref{Eq:STVC-vector} as:
\begin{equation} \label{Eq:STVC-decompose}
\boldsymbol{y}=\tilde{\bd{X}}\operatorname{vec}\left(\boldsymbol{W}(\boldsymbol{V}\odot\boldsymbol{U})^{\top}\right)+\boldsymbol{\epsilon},
\end{equation}
where $\tilde{\bd{X}}=\left(\boldsymbol{I}_{MN}\odot\boldsymbol{X}_{(3)}\right)^{\top}$ denotes the expanded covariate matrix. The number of parameters in \eqref{Eq:STVC-decompose} is $R(M+N+P)$, which is substantially less than $M N P$ in \eqref{Eq:STVC-vector}.

\begin{figure}[!t]
\centering
\includegraphics[width=0.9\textwidth]{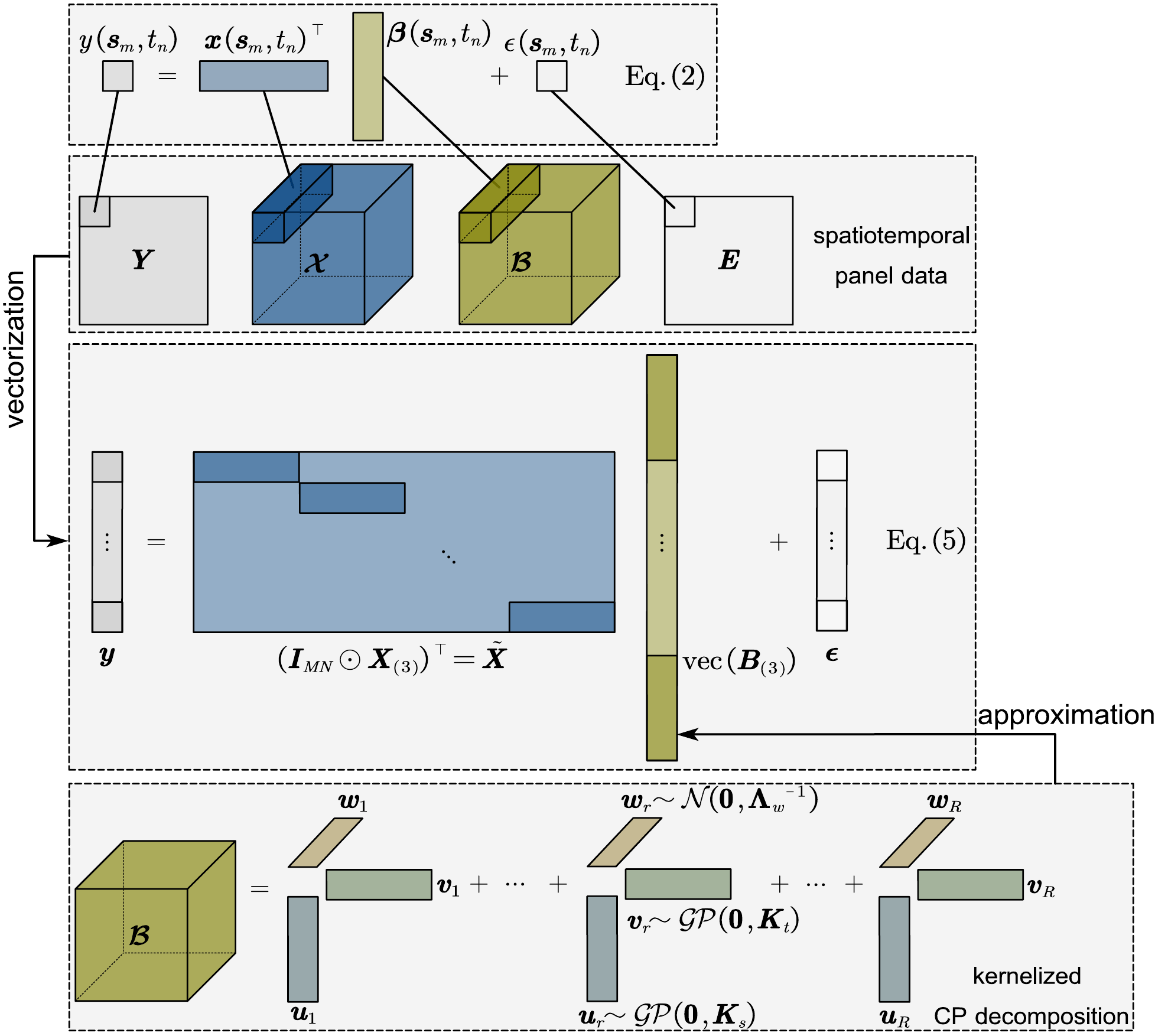}
\caption{Illustration of the proposed BKTR framework.}
\label{fig:illustration-BKTL}
\end{figure}

\begin{figure}[!t]
\centering
\includegraphics[width=0.36\textwidth]{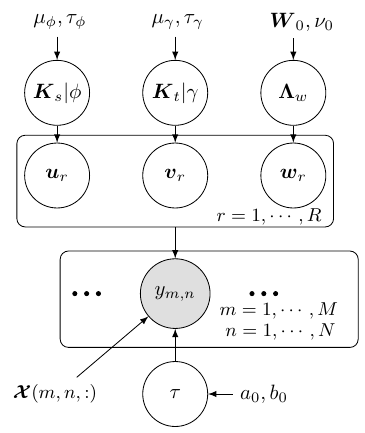}
\caption{Graphical model of BKTR.}
\label{fig:graphical-BKTL}
\end{figure}

Local spatial and temporal processes are critical to the modeling of spatiotemporal data. However, as mentioned above, the low-rank assumption alone cannot encode such local dependencies. To address this issue, we assume specific GP priors on $\bd{U}$ and $\bd{V}$ following the GP factor analysis strategy \citep{luttinen2009variational}, use a conjugate normal prior on $\boldsymbol{W}$, and then develop a fully Bayesian approach to estimate the model in Eq.~\eqref{Eq:STVC-decompose}. Figure~\ref{fig:illustration-BKTL} illustrates the proposed framework, which is referred to as \textit{Bayesian Kernelized Tensor Regression} (BKTR) in the remainder of this paper. The graphical model of BKTR is shown in Figure~\ref{fig:graphical-BKTL}.

As mentioned in the introduction, in real-world applications the dependent data is often partially observed on a set $\Omega$ of observation indices, with $|\Omega|< MN$. This means that we only observe a subset of entries $y_{m,n}$, for $\forall(s_m,t_m) \in \Omega$. We denote by $\boldsymbol{D}\in\mathbb{R}^{M\times N}$ a binary indicator matrix with $d_{m,n}=1$ if $(m,n)\in\Omega$ and $d_{m,n}=0$ otherwise, and by $\boldsymbol{O}$ a binary matrix of $|\Omega|\times MN$ formed by removing the rows corresponding to the zero values in $\operatorname{vec}(\boldsymbol{D})$ from a $MN\times MN$ identity matrix. The vector of observed data can be obtained by
$\boldsymbol{y}_{\Omega}=\boldsymbol{O}\boldsymbol{y}$. Therefore, we have:
\begin{equation} \label{eq:y_omega}
\boldsymbol{y}_{\Omega}\sim\mathcal{N}\left(\boldsymbol{O}\left(\tilde{\bd{X}}\operatorname{vec}\left(\boldsymbol{W}(\boldsymbol{V}\odot\boldsymbol{U})^{\top}\right)\right),\tau^{-1}\boldsymbol{I}_{|\Omega|}\right).
\end{equation}

For spatial and temporal factor matrices $\boldsymbol{U}$ and $\boldsymbol{V}$, we use identical GP priors on the component vectors:
\begin{equation} \label{eq:factorUV}
\begin{aligned}
    \boldsymbol{u}_{r}&\sim\mathcal{GP}\left(\boldsymbol{0},\boldsymbol{K}_{s}\right),~r=1,\ldots,R, \\
    \boldsymbol{v}_{r}&\sim\mathcal{GP}\left(\boldsymbol{0},\boldsymbol{K}_{t}\right),~r=1,\ldots,R, \\
\end{aligned}
\end{equation}
where $\boldsymbol{K}_{s}\in\mathbb{R}^{M\times M}$ and $\boldsymbol{K}_{t}\in\mathbb{R}^{N\times N}$ are the spatial and temporal covariance matrices built from two valid kernel functions $k_{s}(\bd{s}_m,\bd{s}_{m^{\prime}};\phi)$ and $k_{t}(t_n,t_{n^{\prime}};\gamma)$, respectively, with $\phi$ and $\gamma$ being kernel length-scale hyperparameters. Note that we capture the variance through $\boldsymbol{W}$ and thus restrict $\boldsymbol{K}_{s}$ and $\boldsymbol{K}_{t}$ to being correlation matrices by setting the variance to one. We reparameterize the kernel hyperparameters as log-transformed variables to ensure their positivity and assume normal priors on them, i.e., $\log(\phi)\sim\mathcal{N}(\mu_{\phi},\tau_{\phi}^{-1}),~\log(\gamma)\sim\mathcal{N}(\mu_{\gamma},\tau_{\gamma}^{-1})$.
For the factor matrix $\boldsymbol{W}$, we assume all columns follow an identical zero-mean Gaussian distribution with a conjugate Wishart prior on the precision matrix:
\begin{equation} \label{eq:factorW}
\begin{aligned}
    \boldsymbol{w}_{r} &\sim\mathcal{N}\left(\boldsymbol{0},\boldsymbol{\Lambda}_{w}^{-1}\right),~r=1,\ldots,R, \\
    \boldsymbol{\Lambda}_{w} &\sim\mathcal{W}\left(\boldsymbol{\Psi}_{0},\nu_{0}\right),
    \end{aligned}
\end{equation}
where $\boldsymbol{\Psi}_{0}$ is a $P\times P$ positive-definite scale matrix and $\nu_{0}>P-1$ denotes the degrees of freedom. Finally, we use a conjugate Gamma prior $\tau\sim\text{Gamma}(a_{0},b_{0})$ on the noise precision $\tau$ defined in Eq.~(\ref{eq:y_omega}).

Based on the assumed priors and hyperpriors, we can write the covariance function of the coefficients $\boldsymbol{\mathcal{B}}$ modeled with BKTR. Specifically, consider a data pair in the input space with the indices $\left(m,n,p\right)$ and $\left(m',n',p'\right)$, the covariance between the two entries is
\begin{equation} \label{Eq:CovB}
\operatorname{Cov}\left(\boldsymbol{\mathcal{B}}\left(m,n,p\right),\boldsymbol{\mathcal{B}}\left(m',n',p'\right)\right)=\operatorname{Cov}\left(\sum_{r=1}^{R}\boldsymbol{u}_r\left(m\right)\boldsymbol{v}_r\left(n\right)\boldsymbol{w}_r\left(p\right),\sum_{r'=1}^{R}\boldsymbol{u}_{r'}\left(m'\right)\boldsymbol{v}_{r'}\left(n'\right)\boldsymbol{w}_{r'}\left(p'\right)\right).
\end{equation}
Given the prior on $\boldsymbol{w}_r$ (see Eq.~\eqref{eq:factorW}), we have
\begin{equation} \label{Eq:CovW}
\begin{cases}
\operatorname{Cov}\left(\boldsymbol{w}_r\left(p\right),\boldsymbol{w}_{r'}\left(p'\right)\right)=0,~\forall r\neq r', ~\forall p, p' ~(\text{columns are independent})  \\
\operatorname{Cov}\left(\boldsymbol{w}_r\left(p\right),\boldsymbol{w}_{r}\left(p'\right)\right)=\boldsymbol{\Lambda}_w^{-1}\left(p,p'\right).
\end{cases}
\end{equation}
Integrating out $\boldsymbol{w}_r$ in Eq.~\eqref{Eq:CPdecompB} using Eqs.~\eqref{Eq:CovB} and~\eqref{Eq:CovW}, we have
\begin{equation}\label{eq:cov1}
\operatorname{vec}\left(\boldsymbol{B}_{(3)}\right)\mid\boldsymbol{U},\boldsymbol{V},\boldsymbol{\Lambda}_{w}\sim\mathcal{N}\left(\boldsymbol{0},\left(\sum_{r=1}^{R}\left(\boldsymbol{v}_r\boldsymbol{v}_r^{\top}\right)\otimes\left(\boldsymbol{u}_r\boldsymbol{u}_r^{\top}\right)\right)\otimes\boldsymbol{\Lambda}_w^{-1}\right).
\end{equation}
Similarly, if we marginalize $\boldsymbol{U}$ in Eq.~\eqref{Eq:CPdecompB}, we can derive
\begin{equation}\label{eq:cov2}
\left.\operatorname{vec}\left(\boldsymbol{B}_{(1)}\right)\right|\boldsymbol{V},\boldsymbol{W},\boldsymbol{K}_s\sim\mathcal{N}\left(\boldsymbol{0},\left(\sum_{r=1}^{R}\left(\boldsymbol{w}_r\boldsymbol{w}_r^{\top}\right)\otimes\left(\boldsymbol{v}_r\boldsymbol{v}_r^{\top}\right)\right)\otimes\boldsymbol{K}_s\right).
\end{equation}

With the analysis of the variance-covariance matrix in Eqs.~\eqref{eq:cov1} and \eqref{eq:cov2}, we can interpret the kernelized tensor CP factorization as a higher-order extension of the intrinsic model of coregionalization \citep{bonilla2007multi,banerjee2014hierarchical,alvarez2012kernels}, where the coregionalization matrix also has a low-rank specification. However, for model inference we do not use the covariance structure; instead, we directly use tensor factorization to learn the latent factor matrices for fast inference.

\subsection{Model inference}
We use Gibbs sampling to estimate the model parameters, including coefficient factors $\{\boldsymbol{U},\boldsymbol{V},\boldsymbol{W}\}$, the precision $\tau$, and the precision matrix $\boldsymbol{\Lambda}_{w}$. For the kernel hyperparameters $\{\phi,\gamma\}$ whose conditional distributions are not easy to sample from, we use the slice sampler.

\subsubsection{Sampling the coefficient factor matrices \texorpdfstring{$\{\boldsymbol{U},\boldsymbol{V},\boldsymbol{W}\}$}{TEXT}}

Sampling the factor matrices can be considered as a Bayesian linear regression problem. Taking $\bd{W}$ as an example, we can rewrite Eq.~\eqref{eq:y_omega} as:
\begin{equation} \label{Eq:STVC-sampleW}
    \boldsymbol{y}_{\Omega}\sim \mathcal{N}\left(\boldsymbol{O}\left(\tilde{\bd{X}}\left((\boldsymbol{V}\odot\boldsymbol{U})\otimes\boldsymbol{I}_{P}\right)\operatorname{vec}(\boldsymbol{W})\right),\tau^{-1}\boldsymbol{I}_{|\Omega|}\right),
\end{equation}
where $\bd{U},\bd{V}$ are known and $\operatorname{vec}(\boldsymbol{W})$ is the coefficient to estimate. Considering that the priors of each component vector $\boldsymbol{w}_{r}$ are independent and identical, the prior distribution over the whole vectorized $\boldsymbol{W}$ becomes $\operatorname{vec}(\boldsymbol{W})\sim\mathcal{N}(\boldsymbol{0},\boldsymbol{I}_{R}\otimes\boldsymbol{\Lambda}_{w}^{-1})$. Since both likelihood and prior of $\operatorname{vec}(\boldsymbol{W})$ follow Gaussian distributions, its posterior is also Gaussian with mean $\boldsymbol{\mu}_{W}^{*}$ and precision $\boldsymbol{\Lambda}_{W}^{*}$, such as:
\begin{equation}
\begin{aligned}
\boldsymbol{\Lambda}_{W}^{*}&=\tau\boldsymbol{H}_{W}^{\top}\boldsymbol{H}_{W}+\boldsymbol{I}_{R}\otimes\boldsymbol{\Lambda}_{w},~\boldsymbol{\mu}_{W}^{*}=\tau(\boldsymbol{\Lambda}_{W}^{*})^{-1}\left(\boldsymbol{H}_{W}^{\top}\boldsymbol{y}_{\Omega}\right),
\end{aligned}
\end{equation}
where $\boldsymbol{H}_{W}=\boldsymbol{O}\left(\tilde{\bd{X}}\left((\boldsymbol{V}\odot\boldsymbol{U})\otimes\boldsymbol{I}_{P}\right)\right)$ with size $|\Omega|\times RP$.
Sampling from $\mathcal{N}\left(\boldsymbol{\mu}_{W}^{*},\boldsymbol{\Lambda}_{W}^{*}\right)$ is mainly dominated by the Cholesky decomposition of $\boldsymbol{\Lambda}_{W}^{*}$, and the procedure requires $\mathcal{O}(R^2P^2)$ in storage and $\mathcal{O}(R^3P^3)$ in time.
The posterior distributions of $\boldsymbol{U}$ and $\boldsymbol{V}$ can be obtained similarly using different tensor unfoldings. In order to sample $\boldsymbol{U}$, we use the mode-1 unfolding in Eq.~\eqref{Eq:unfolding-CP} and reconstruct the regression model with $\operatorname{vec}(\bd{U})$ as coefficients:
\begin{equation} \label{Eq:YonU}
\boldsymbol{y}_{\Omega}=\boldsymbol{O}\left(\tilde{\bd{X}}_{U}\left((\boldsymbol{W}\odot\boldsymbol{V})\otimes\boldsymbol{I}_{M}\right)\operatorname{vec}(\boldsymbol{U})\right)+\boldsymbol{\epsilon}_{\Omega},
\end{equation}
where {{$\tilde{\bd{X}}_{U}=\left(\boldsymbol{X}_{(3)}\odot\boldsymbol{I}_{MN}\right)^{\top}\in\mathbb{R}^{MN\times MNP}$}} and  $\boldsymbol{\epsilon}_{\Omega}\sim\mathcal{N}\left(\boldsymbol{0},\tau^{-1}\boldsymbol{I}_{|\Omega|}\right)$. Thus, the posterior of $\operatorname{vec}(\boldsymbol{U})$ has a closed form---a Gaussian distribution with mean $\boldsymbol{\mu}_{U}^{*}$ and precision $\boldsymbol{\Lambda}_{U}^{*}$, where
\begin{equation}\label{eq:update_U}
\begin{aligned}
\boldsymbol{\Lambda}_{U}^{*}&=\tau\boldsymbol{H}_{U}^{\top}\boldsymbol{H}_{U}+\boldsymbol{K}_{U}^{-1},~\boldsymbol{\mu}_{U}^{*}=\tau(\boldsymbol{\Lambda}_{U}^{*})^{-1}\left(\boldsymbol{H}_{U}^{\top}\boldsymbol{y}_{\Omega}\right),
\end{aligned}
\end{equation}
with $\boldsymbol{K}_{U}=\boldsymbol{I}_{R}\otimes\boldsymbol{K}_{s}$ and $\boldsymbol{H}_{U}=\boldsymbol{O}\left(\tilde{\bd{X}}_{U}\left((\boldsymbol{W}\odot\boldsymbol{V})\otimes\boldsymbol{I}_{M}\right)\right)\in\mathbb{R}^{|\Omega|\times MR}$. The posterior for $\operatorname{vec}(\boldsymbol{V})$ can be obtained by applying the mode-2 tensor unfolding. It should be noted that the above derivation provides the posterior for the whole factor matrix, i.e., $\left\{\operatorname{vec}(\bd{U}),\operatorname{vec}(\bd{V}),\operatorname{vec}(\bd{W})\right\}$, so the time complexity is $\mathcal{O}(R^3(M^3+N^3+P^3))$. We can further reduce the computational cost by sampling $\left\{\bd{u}_r,\boldsymbol{v}_{r},\boldsymbol{w}_{r}\right\}$ one by one for $r=1,\ldots,R$ as in \citet{luttinen2009variational}. In this case, the time complexity in learning these factor matrices can be further reduced to $\mathcal{O}(R(M^3+N^3+P^3))$ at the cost of slow/poor mixing.

\subsubsection{Sampling kernel hyperparameters \texorpdfstring{$\{\phi,\gamma\}$}{TEXT}}
As shown in Figure~\ref{fig:graphical-BKTL}, sampling kernel hyperparameters conditional on the factor matrices should be straightforward through the Metropolis-Hastings algorithm. However, in practice, conditioning on the latent variables $\{\boldsymbol{U},\boldsymbol{V}\}$ in such hierarchical GP models usually induces sharply peaked conditional posteriors over $\{\phi,\gamma\}$, making the Markov chains mix slowly and resulting in poor updates \citep{murray2010slice}. To address this issue, we integrate out the latent factors from the model to get the marginal likelihood, and sample $\phi$ and $\gamma$ from their marginal posterior distributions based on the slice sampling approach \citep{neal2003slice}; i.e., we integrate out $\boldsymbol{U}$ when deriving the marginal posterior distribution $p\left(\phi\given \boldsymbol{y}_{\Omega},\boldsymbol{V},\boldsymbol{W},\tau,\boldsymbol{\mathcal{X}}\right)$, and likewise we build the marginal posterior $p\left(\gamma\given \boldsymbol{y}_{\Omega},\boldsymbol{U},\boldsymbol{W},\tau,\boldsymbol{\mathcal{X}}\right)$ by integrating out $\boldsymbol{V}$. Compared with recent related studies that sample hyperparameters conditioned on latent factors, e.g., \citet{zhang2022spatial}, the marginalization of latent factors can obtain tractable posteriors and avoid the possible issues in MCMC convergence. Note that in the work of \citet{murray2010slice}, the proposed auxiliary variable strategy can also be used to handle outcomes with distributions beyond Gaussian. In addition, for kernel functions that contain more than one hyperparameter, one can apply a modified rectangle slice sampler \citep{neal2003slice}.

Let us consider for example the hyperparameter of $\boldsymbol{K}_s$, i.e., $\phi$. As $\operatorname{vec}(\boldsymbol{U})\sim\mathcal{N}(\boldsymbol{0},\boldsymbol{K}_{U})$ with  $\boldsymbol{K}_{U}=\boldsymbol{I}_{R}\otimes\boldsymbol{K}_s$, we integrate out $\operatorname{vec}(\boldsymbol{U})$ in Eq.~\eqref{Eq:YonU} and obtain:
\begin{equation}\label{eq:marginal_likeli}\log{p\left(\boldsymbol{y}_{\Omega}\given\phi,\boldsymbol{V},\boldsymbol{W},\tau,\boldsymbol{\mathcal{X}}\right)}=-\frac{1}{2}\boldsymbol{y}_{\Omega}^{\top}\boldsymbol{K}_{\left.\boldsymbol{y}\right|\phi}^{-1}\boldsymbol{y}_{\Omega}-\frac{1}{2}\log{\left|\boldsymbol{K}_{\left.\boldsymbol{y}\right|\phi}\right|}-\frac{|\Omega|}{2}\log{2\pi},
\end{equation}
where $\boldsymbol{K}_{\left.\boldsymbol{y}\right|\phi}=\boldsymbol{H}_{U}\boldsymbol{K}_{U}\boldsymbol{H}_{U}^{\top}+\tau^{-1}\boldsymbol{I}_{|\Omega|}\in\mathbb{R}^{|\Omega|\times|\Omega|}$ and $\boldsymbol{H}_{U}$ is the same as the definition used in Eq.~\eqref{eq:update_U}, i.e., $\boldsymbol{H}_{U}=\boldsymbol{O}\left(\tilde{\bd{X}}_{U}\left((\boldsymbol{W}\odot\boldsymbol{V})\otimes\boldsymbol{I}_{M}\right)\right)$.
The marginal posterior of $\phi$ becomes:
\begin{equation} \label{Eq:Marginal}
\begin{aligned}
\log{p\left(\phi\given\boldsymbol{y}_{\Omega},\boldsymbol{V},\boldsymbol{W},\tau,\boldsymbol{\mathcal{X}}\right)}\propto&\log{p(\phi)}-\frac{1}{2}\boldsymbol{y}_{\Omega}^{\top}\boldsymbol{K}_{\left.\boldsymbol{y}\right|\phi}^{-1}\boldsymbol{y}_{\Omega}-\frac{1}{2}\log{\left|\boldsymbol{K}_{\left.\boldsymbol{y}\right|\phi}\right|} \\
\propto&\log{p(\phi)}+\frac{1}{2}\tau^2\boldsymbol{y}_{\Omega}^{\top}\boldsymbol{H}_{U}\left(\boldsymbol{I}_{R}\otimes\boldsymbol{K}_{s}^{-1}+\tau\boldsymbol{H}_{U}^{\top}\boldsymbol{H}_{U}\right)^{-1}\boldsymbol{H}_{U}^{\top}\boldsymbol{y}_{\Omega} \\
&-\frac{1}{2}\log{\left|\boldsymbol{I}_{R}\otimes\boldsymbol{K}_{s}^{-1}+\tau\boldsymbol{H}_{U}^{\top}\boldsymbol{H}_{U}\right|}-\frac{R}{2}\log{\left|\boldsymbol{K}_{s}\right|},
\end{aligned}
\end{equation}
where we compute $\boldsymbol{y}_{\Omega}^{\top}\boldsymbol{K}_{\left.\boldsymbol{y}\right|\phi}^{-1}\boldsymbol{y}_{\Omega}$ based on the Woodbury matrix identity, and use the matrix determinant lemma to compute $\log{\left|\boldsymbol{K}_{\left.\boldsymbol{y}\right|\phi}\right|}$. The detailed derivation is given in Appendix~\ref{appA}. Computing Eq.~\eqref{Eq:Marginal} involves the Cholesky factorization of $\boldsymbol{L}=\texttt{chol}(\boldsymbol{I}_{R}\otimes\boldsymbol{K}_{s}^{-1}+\tau\boldsymbol{H}_{U}^{\top}\boldsymbol{H}_{U})$ of size $MR\times MR$; therefore, the overall time complexity is $\mathcal{O}(M^3R^3)$.

The slice sampling approach is robust to the selection of the sampling scale and easy to implement. Sampling $\gamma$ can be achieved in a similar way. Note that, as mentioned, the sampling is performed on the log-transformed variables to avoid numerical issues. The detailed sampling process for kernel hyperparameters is provided in Appendix~\ref{appB}.
We can also introduce different kernel functions (or the same kernel function with different hyperparameters) for each factor vector $\bd{u}_r$ and $\boldsymbol{v}_{r}$ as in \citet{luttinen2009variational}. In this case, the marginal posterior of the kernel hyperparameters can be derived in a similar way as in Eq.~\eqref{Eq:Marginal}.

\subsubsection{Sampling \texorpdfstring{$\boldsymbol{\Lambda}_{w}$}{TEXT}}
Given the conjugate Wishart prior, the posterior distribution of $\boldsymbol{\Lambda}_{w}$ is $\left.\boldsymbol{\Lambda}_{w}\right|\boldsymbol{W},\boldsymbol{\Psi}_{0},\nu_{0}\sim\mathcal{W}\left(\boldsymbol{\Psi}^{*},\nu^{*}\right)$,
where $[\boldsymbol{\Psi}^{*}]^{-1}=\boldsymbol{W}\boldsymbol{W}^{\top}+\boldsymbol{\Psi}_{0}^{-1}$ and $\nu^{*}=\nu_{0}+R$.

\subsubsection{Sampling the precision \texorpdfstring{$\tau$}{TEXT}}
Since we used a conjugate Gamma prior, the posterior distribution of $\tau$ is also a Gamma distribution with shape ($a^{\ast}$) and rate ($b^{\ast}$) being
\begin{equation}
    {a^{*}=a_{0}+\frac{1}{2}|\Omega|},~{b^{*}=b_{0}+\frac{1}{2}\left\|\boldsymbol{y}_{\Omega}-\boldsymbol{O}\left(\tilde{\bd{X}}\operatorname{vec}\left(\boldsymbol{W}(\boldsymbol{V}\odot\boldsymbol{U})^{\top}\right)\right)\right\|_{2}^{2}}.
\end{equation}

\begin{algorithm}[!t]
\caption{MCMC sampling process of $\operatorname{BKTR}(\boldsymbol{y}_{\Omega},\boldsymbol{\mathcal{X}},R,K_{1},K_{2})$}
\label{alg:BKTL}
{Initialize $\{{\boldsymbol{U},\boldsymbol{V},\boldsymbol{W}}\}$ as normally distributed random values, $\phi=\gamma=1$, and ${\boldsymbol{\Lambda}_{w}\sim\mathcal{W}(\boldsymbol{I}_{P},P)}$. Set $\mu_{\phi}=\mu_{\gamma}=\log (1)$, $\tau_{\phi}=\tau_{\gamma}=10$, and $a_0=b_0=10^{-4}$.
} \\
\For{$k=1:K_{1}+K_{2}$}{
Sample kernel hyperparameter $\phi$ using the Algorithm in Appendix~\ref{appB}; \\

Sample factor $\operatorname{vec}(\boldsymbol{U})$ from a Gaussian distribution: $\left.\operatorname{vec}(\boldsymbol{U})\right|-\sim\mathcal{N}\left(\boldsymbol{\mu}_{U}^{*},\left(\boldsymbol{\Lambda}_{U}^{*}\right)^{-1}\right)$, $\boldsymbol{\mu}_{U}^{*}=\tau(\boldsymbol{\Lambda}_{U}^{*})^{-1}\boldsymbol{H}_{U}^{\top}\boldsymbol{y}_{\Omega}$, $\boldsymbol{\Lambda}_{U}^{*}=\tau\boldsymbol{H}_{U}^{\top}\boldsymbol{H}_{U}+\boldsymbol{I}_{R}\otimes\boldsymbol{K}_{s}^{-1}$, $\boldsymbol{H}_{U}=\boldsymbol{O}\left(\left(\boldsymbol{X}_{(3)}\odot\boldsymbol{I}_{MN}\right)^{\top}\left((\boldsymbol{W}\odot\boldsymbol{V})\otimes\boldsymbol{I}_{M}\right)\right)$; \\

Sample kernel hyperparameter $\gamma$ using the Algorithm in Appendix~\ref{appB}; \\

Sample factor $\operatorname{vec}(\boldsymbol{V})$ from a Gaussian distribution: $\left.\operatorname{vec}(\boldsymbol{V})\right|-\sim\mathcal{N}\left(\boldsymbol{\mu}_{V}^{*},\left(\boldsymbol{\Lambda}_{V}^{*}\right)^{-1}\right)$, $\boldsymbol{\mu}_{V}^{*}=\tau(\boldsymbol{\Lambda}_{V}^{*})^{-1}\boldsymbol{H}_{V}^{\top}\boldsymbol{y}_{\Omega}^{\top}$, $\boldsymbol{\Lambda}_{V}^{*}=\tau\boldsymbol{H}_{V}^{\top}\boldsymbol{H}_{V}+\boldsymbol{I}_{R}\otimes\boldsymbol{K}_{t}^{-1}$, $\boldsymbol{H}_{V}=\boldsymbol{O}^{\prime}\left(\left(\boldsymbol{X}_{(3)}^{\top}\odot\boldsymbol{I}_{MN}\right)^{\top}\left((\boldsymbol{W}\odot\boldsymbol{U})\otimes\boldsymbol{I}_{N}\right)\right)$, where $\boldsymbol{O}^{\prime}\in\mathbb{R}^{|\Omega|\times MN}$ is a matrix removing the rows corresponding to the zeros in $\operatorname{vec}(\boldsymbol{D}^{\top})$ from $\boldsymbol{I}_{MN}$, $\boldsymbol{X}_{(3)}^{\top}$ is the mode-3 unfolding of $\boldsymbol{\mathcal{X}}^{\top}\in\mathbb{R}^{N\times M\times P}$, whose frontal slices are the transpose matrices of frontal slices of $\boldsymbol{\mathcal{X}}$, and $\boldsymbol{y}_{\Omega}^{\top}=\boldsymbol{O}^{\prime}\operatorname{vec}\left(\boldsymbol{Y}^{\top}\right)$; \\

Sample hyperparameters $\boldsymbol{\Lambda}_{w}$ from a Wishart distribution: $\left.\boldsymbol{\Lambda}_{w}\right|-\sim\mathcal{W}\left(\left(\boldsymbol{W}\boldsymbol{W}^{\top}+\boldsymbol{I}_{P}^{-1}\right)^{-1},P+R\right)$; \\

Sample factor $\operatorname{vec}(\boldsymbol{W})$ from a Gaussian distribution: $\left.\operatorname{vec}(\boldsymbol{W})\right|-\sim\mathcal{N}\left(\boldsymbol{\mu}_{W}^{*},\left(\boldsymbol{\Lambda}_{W}^{*}\right)^{-1}\right)$, $\boldsymbol{\mu}_{W}^{*}=\tau(\boldsymbol{\Lambda}_{W}^{*})^{-1}\boldsymbol{H}_{W}^{\top}\boldsymbol{y}_{\Omega}$, $\boldsymbol{\Lambda}_{W}^{*}=\tau\boldsymbol{H}_{W}^{\top}\boldsymbol{H}_{W}+\boldsymbol{I}_{R}\otimes\boldsymbol{\Lambda}_{w}$, $\boldsymbol{H}_{W}=\boldsymbol{O}\left(\left(\boldsymbol{I}_{MN}\odot\boldsymbol{X}_{(3)}\right)^{\top}\left((\boldsymbol{V}\odot\boldsymbol{U})\otimes\boldsymbol{I}_{P}\right)\right)$; \\

Sample precision $\tau$ from a Gamma distribution: $\left.\tau\right|-\sim\text{Gamma}\left(a_{0}+\frac{1}{2}|\Omega|,b_{0}+\frac{1}{2}\left\|\boldsymbol{y}_{\Omega}-\boldsymbol{O}\left(\tilde{\bd{X}}\operatorname{vec}\left(\boldsymbol{W}(\boldsymbol{V}\odot\boldsymbol{U})^{\top}\right)\right)\right\|_{2}^{2}\right)$; \\

\If{$k>K_{1}$}{
{Collect the sample} ${\boldsymbol{U}^{(k-K_1)}=\bd{U},\boldsymbol{V}^{(k-K_1)}=\bd{V},\boldsymbol{W}^{(k-K_1)}=\bd{W}}$;\\

Compute $\boldsymbol{\mathcal{B}}^{(k-K_1)}$ using $\boldsymbol{\mathcal{B}}=\sum_{r=1}^{R}\boldsymbol{u}_{r}\circ\boldsymbol{v}_{r}\circ\boldsymbol{w}_{r}$.}
}
\Return $\{\boldsymbol{\mathcal{B}}^{(k)}\}_{k=1}^{K_2}$ to approximate posterior coefficients and estimate unobserved data.
\end{algorithm}

\subsection{Model implementation}
We summarize the implementation of BKTR in Algorithm~\ref{alg:BKTL}. It should be noted that we update the correlated latent factors and the corresponding hyperparameters as one block in the Gibbs sampler to further ensure model convergence \citep{knorr2002block}. Specifically, we take $\left\{\phi,\boldsymbol{U}\right\}$ as a block and update $\phi$ from its marginal posterior with a slice sampler followed by sampling $\boldsymbol{U}$ from $p\left(\boldsymbol{U}\given\phi,-\right)$, and then similarly sample $\left\{\gamma,\boldsymbol{V}\right\}$ as another block. For MCMC inference, we run $K_1$ iterations as burn-in and take the following $K_2$ samples for estimation.

\subsection{Model scalability} \label{sec_scalability}
Compared to the original STVC approach \citep{gelfand2003spatial}, which requires $\mathcal{O}\left(|\Omega|^3\right)$ for hyperparameter sampling and $\mathcal{O}\left(|\Omega|^3P^3\right)$ for coefficient variable learning. BKTR reduces the cost of updating hyperparameters and coefficients to  $\mathcal{O}\left(R^3\left(M^3+N^3+P^3\right)\right)$; e.g., updating the factor matrix $\boldsymbol{U}$ following Eq.~\eqref{eq:update_U} requires $\mathcal{O}(M^3R^3)$ in time. For the slice sampling of kernel hyperparameters $\phi$, each update inside the slice sampling loop (see Algorithm~\ref{alg:HyperSample}) requires $\mathcal{O}(M^3R^3)$ in computing the likelihood in Eq.~\eqref{Eq:Marginal}. Such substantial gains in computing time allow us to analyze large-scale real-world spatiotemporal data and multi-dimensional relations, where generally STVC is infeasible.

Given the above analysis, the computation cost BKTR depends on rank $R$ and BKTR can work on large data sets such as $R\times M\approx 10^3\sim 10^4$ (the same applies to $N$ and $P$). However, it should be noted that the default BKTR still encounters computational issues when the number of spatial locations $M$ (or time points $N$) becomes very large (e.g., say $M>10^4$). We next discuss several solutions when $M$ becomes large. Following \citet{luttinen2009variational}, the most straightforward solution is to update the three factor matrices column by column, which reduces the time cost to $\mathcal{O}\left(R\left(M^3+N^3+P^3\right)\right)$. Through this updating scheme, the Kroneceker products for learning latent factors and hyperparameters can be avoided. Taking $\boldsymbol{u}_r$ (i.e., the $r$-th column in $\boldsymbol{U}$) as an example, the posterior of $\boldsymbol{u}_r$ when updated independently still follows a Gaussian distribution $\mathcal{N}\left(\boldsymbol{\mu}_{u_r}^{\ast},\left(\boldsymbol{\Lambda}_{u_r}^{\ast}\right)^{-1}\right)$, where
\begin{equation}
\boldsymbol{\Lambda}_{u_r}^{\ast}=\tau\boldsymbol{H}_{u_r}^{\top}\boldsymbol{H}_{u_r}+\boldsymbol{K}_s^{-1}, \ \boldsymbol{\mu}_{u_r}^{\ast}=\tau\left(\boldsymbol{\Lambda}_{u_r}^{\ast}\right)^{-1}\left(\boldsymbol{H}_{u_r}^{\top}\boldsymbol{y}_r\right)
\end{equation} with $\boldsymbol{H}_{u_r}=\boldsymbol{O}\left(\tilde{\boldsymbol{X}}_{U}\left(\left(\boldsymbol{w}_{r}\otimes\boldsymbol{v}_r\right)\otimes\boldsymbol{I}_{M}\right)\right)\in\mathbb{R}^{|\Omega|\times M}$ and $\boldsymbol{y}_r=\boldsymbol{O}\big(\boldsymbol{y}-\tilde{\boldsymbol{X}}_{U}\big(\left(\boldsymbol{W}_{-r}\odot\boldsymbol{V}_{-r}\right)\\\otimes\boldsymbol{I}_{M}\big)\operatorname{vec}\left(\boldsymbol{U}_{-r}\right)\big)\in\mathbb{R}^{|\Omega|}$. $\left\{\boldsymbol{U}_{-r},\boldsymbol{V}_{-r},\boldsymbol{W}_{-r}\right\}$ are the factor matrices without the $r$th columns and $\boldsymbol{H}_{u_r}^\top\boldsymbol{H}_{u_r}\in\mathbb{R}^{M\times M}$ is a diagonal matrix. \citet{luttinen2009variational} also suggested to use sparse approximation and predictive processes \citep{banerjee2008gaussian,titsias2009variational} to model the each latent factor vector when $M$ becomes large, i.e., using $M_0$ inducing points where $M_0 \ll M$. This reduces the time complexity to $\mathcal{O}(RM_0^2M)$.

Given that $\boldsymbol{H}_{u_r}^{\top}\boldsymbol{H}_{u_r}$ becomes an $M\times M$ diagonal matrix, if we replace the Gaussian prior with a Gaussian Markov random field (GRMF) \citep{rue2005gaussian} for which the precision matrix $\boldsymbol{K}_s^{-1}$ becomes a sparse banded matrix, we can then use sparse matrix algorithms (as $\boldsymbol{\Lambda}_{u_r}^{\ast}$ also becomes a sparse banded matrix) to model large-scale data sets with $M\approx 10^4\sim 10^6$. Similarly, NNGP can also be used here to model the spatial latent factor $\boldsymbol{U}$ for scalable inference \citep{finley2019efficient}. The column-by-column approach also offers the flexibility to learn different kernel hyperparameter $\phi_r$ (and thus covariance $\boldsymbol{K}_s^r$) for each latent factor vector $\boldsymbol{u}_r$. The marginal likelihood conditioning on $\phi_r$ in Eq.~\eqref{eq:marginal_likeli} becomes $\log p\left(\boldsymbol{y}_{\Omega}\given\phi_r,-\right)\propto-\frac{1}{2}\boldsymbol{y}_{r}^{\top}\boldsymbol{K}_{\left.\boldsymbol{y}_r\right|\phi}^{-1}\boldsymbol{y}_{r}-\frac{1}{2}\log\left|\boldsymbol{K}_{\left.\boldsymbol{y}_r\right|\phi}\right|$, where $\boldsymbol{K}_{\left.\boldsymbol{y}_r\right|\phi}=\boldsymbol{H}_{u_r}\boldsymbol{K}_s^r\boldsymbol{H}_{u_r}^{\top}+\tau^{-1}\boldsymbol{I}_{|\Omega|}\in\mathbb{R}^{|\Omega|\times|\Omega|}$. Correspondingly, the marginal posterior in Eq.~\eqref{Eq:Marginal} becomes:
\begin{equation}
\begin{aligned}
&\log p\left(\phi_r\given\boldsymbol{y}_{\Omega},\boldsymbol{U}_{-r},\boldsymbol{V},\boldsymbol{W},\tau,\boldsymbol{\mathcal{X}}\right)\propto\log p\left(\phi_r\right)-\frac{1}{2}\log\left|\boldsymbol{K}_s^r\right|+\\
&\frac{1}{2}\tau^2\boldsymbol{y}_r^{\top}\boldsymbol{H}_{u_r}\left((\boldsymbol{K}_s^r)^{-1}+\tau\boldsymbol{H}_{u_r}^{\top}\boldsymbol{H}_{u_r}\right)^{-1}\boldsymbol{H}_{u_r}^{\top}\boldsymbol{y}_r-\frac{1}{2}\log\left|(\boldsymbol{K}_s^r)^{-1}+\tau\boldsymbol{H}_{u_r}^{\top}\boldsymbol{H}_{u_r}\right|.
\end{aligned}
\end{equation}

Lastly, considering that learning hyperparameters via slice sampling is expensive, another possible solution to further reduce the computational cost is to use cross-validation to specify kernel hyperparameters as in \citet{finley2019efficient}. {The choice of kernel hyperparameters is often not that sensitive in regression and thus cross-validation can be performed at a moderately crude resolution. However, it could still be time-consuming to directly implement the cross-validatory approach based on MCMC sampling, since such an iterative process needs to be run several folds/times for each combination of hyperparameters.}

\section{Simulation Study}
\label{sec:simulation}
In this section, we evaluate the performance of BKTR on three simulated data sets. All simulations are performed on a laptop with a 6-core Intel Xenon 2.60 GHz CPU and 32GB RAM. Specifically, we conduct three studies: (1) a low-rank structured association analysis to test the estimation accuracy and statistical properties of BKTR with different rank settings and in scenarios with different observation/missing rates, (2) a small-scale analysis to compare BKTR with STVC and a pure low-rank tensor regression model, and (3) a moderately sized analysis to test the performance of BKTR on more practical STVC modeling.

\subsection{Simulation 1: performance and statistical properties of BKTR}
\subsubsection{Simulation setting}
To fully evaluate the properties of BKTR, we first simulate a data set where $\boldsymbol{\mathcal{B}}$ is constructed following an exact CP decomposition. We generate $M=300$ spatial locations in a $[0,10]\times[0,10]$ square and $N=100$ evenly located time points in $[0,10]$. The variable ${\boldsymbol{\mathcal{X}}\in\mathbb{R}^{300\times100\times5}}$ ${(P=5)}$ contains an intercept and four spatiotemporal covariates as follows:
\begin{equation} \label{Eq:X_generate}
\begin{aligned}
    \bd{\cal{X}}&(:,:,p=1)=\operatorname{ones}(M,N),~~~~~~~~~~~~~~~~~~~~~~~~~~~~~~~\text{(for the intercept)}\\
    \boldsymbol{\mathcal{X}}&(:,:,p=2,3)=\boldsymbol{1}_{N}^{\top}\otimes\boldsymbol{x}_{s}^{p},\text{with } \boldsymbol{x}_{s}^{p}\sim\mathcal{N}(\boldsymbol{0},\boldsymbol{I}_{M}),~~~~\text{(covariates only vary with space)}\\
    \boldsymbol{\mathcal{X}}&(:,:,p=4,5)=\boldsymbol{1}_{M}\otimes(\boldsymbol{x}_{t}^{p})^{\top},\text{with }\boldsymbol{x}_{t}^{p}\sim\mathcal{N}(\boldsymbol{0},\boldsymbol{I}_{N}).\text{ (covariates only vary with time)}
\end{aligned}
\end{equation}
We set the true CP rank $R=10$ and generate the coefficient variable $\boldsymbol{\mathcal{B}}\in\mathbb{R}^{300\times100\times5}$ as $\boldsymbol{\mathcal{B}}=\sum_{r=1}^{R}\boldsymbol{u}_{r}\circ\boldsymbol{v}_{r}\circ\boldsymbol{w}_{r}$, where $\boldsymbol{u}_{r}\sim\mathcal{N}(\boldsymbol{0},\boldsymbol{K}_{s}),~\boldsymbol{v}_{r}\sim\mathcal{N}(\boldsymbol{0},\boldsymbol{K}_{t}),\text{ and }\boldsymbol{w}_{r}\sim\mathcal{N}(\boldsymbol{0},\boldsymbol{\Lambda}_{w}^{-1})$, for $r=1,\ldots,R$. 
Covariance matrices $\boldsymbol{K}_{s}\in\mathbb{R}^{300\times300}$ and $\boldsymbol{K}_{t}\in\mathbb{R}^{100\times100}$ are computed from a Mat\'ern 3/2 kernel $k_s(\bd{s}_{m},\bd{s}_{m'})=\sigma_{s}^2\left(1+\frac{\sqrt{3}d}{\phi}\right)\exp{\left(-\frac{\sqrt{3}d}{\phi}\right)}$ ($d$ is the distance between locations $\bd{s}_{m}$ and $\bd{s}_{m'}$) and a squared exponential (SE) kernel $k_t(t_{n},t_{n'})=\sigma_{t}^2\exp{\left(-\frac{(t_{n}-t_{n'})^2}{2\gamma^{2}}\right)}$, respectively, with variances $\sigma_s^2=\sigma_t^2=2$, length-scales $\phi=\gamma=1$, and $\boldsymbol{\Lambda}_{w}\sim\mathcal{W}(\boldsymbol{I}_{P},P)$. Based on $\boldsymbol{\mathcal{X}}$ and $\boldsymbol{\mathcal{B}}$, the spatiotemporal data $\boldsymbol{Y}\in\mathbb{R}^{300\times100}$ is then generated following Eq.~(\ref{Eq:STVC-vector}) with $\boldsymbol{\epsilon}\sim\mathcal{N}(\boldsymbol{0},\tau^{-1}\boldsymbol{I}_{MN})$ where $\tau^{-1}=1$.

We test the performance of BKTR under different settings to evaluate its sensitivity to i) the rank specification defined by the user, and ii) the proportion of observed responses, i.e., $\frac{|\Omega|}{MN}$. For i), we specify ${R=\{4,7,10,13,16,20,25,30,35,40\}}$ and randomly sample 50\% of the data $\boldsymbol{Y}$ as observed values. For ii), we applied the model under 5 different observed response rates, where we randomly sample (90\%, 70\%, 50\%, 30\%, 10\%) of $\boldsymbol{Y}$ as observed values and evaluate the performance of BKTR with the true rank setting ($R=10$). In all settings, we assume that the kernel functions are known, i.e., $k_s$ is Mat\'ern 3/2 and $k_t$ is SE. We also include in the analysis a sixth covariate in $\boldsymbol{\mathcal{X}}$ drawn from the standard normal distribution and unrelated to the outcome (the corresponding estimated coefficients should be close to zero). For each setting, we replicate the simulation 40 times and run the MCMC with $K_1=1000$ and $K_2=500$.

In each setting, we measure the estimation accuracy on $\boldsymbol{\mathcal{B}}$ and the prediction accuracy on $\boldsymbol{y}_{\Omega^{c}}$ (unobserved data) using mean absolute error (MAE) and root mean square error (RMSE) between the true values and the corresponding posterior mean estimations, where the true $\boldsymbol{\mathcal{B}}$ values for the random non-significant covariate are set to zero. For evaluating the performance on uncertainty estimation, we assess interval coverage (CVG) \citep{heaton2019case}, interval score (INT) of the 95\% credible intervals (CI), and continuous ranked probability score (CRPS) \citep{gneiting2007strictly} of all $\boldsymbol{\mathcal{B}}$ values. Note that all CIs are obtained based on the $K_2$ samples after burn-in. Detailed definitions of all these performance metrics are given in Appendix~\ref{appC}.

\subsubsection{Results}

\begin{table}[!t]
\small
\centering
\caption{Performance of BKTR ($R=10$) on Simulation 1, with 50\% $\boldsymbol{Y}$ missing.}
\begin{tabular}{ccccc}
\toprule
$\text{MAE}_{\boldsymbol{\mathcal{B}}}$/$\text{RMSE}_{\boldsymbol{\mathcal{B}}}$ & $\text{MAE}_{\boldsymbol{y}_{\Omega^c}}$/$\text{RMSE}_{\boldsymbol{y}_{\Omega^c}}$ & CVG & INT & CRPS \\
\midrule
0.21$\pm$0.02/0.33$\pm$0.04 & 0.91$\pm$0.01/1.15$\pm$0.02 & 94.83\%$\pm$0.75\% & 1.04$\pm$0.10 & 0.15$\pm$0.01 \\
\bottomrule
\end{tabular}
\label{tab:simu1}
\end{table}

\begin{figure}[!t]
\centering
\subfigure[Estimated coefficients at location $m=3$, i.e., $\boldsymbol{\mathcal{B}}(3,:,p=1,2,3,4,5,6)$, where the blue lines and dashed areas are posterior means with 95\% CI, and the black dot lines denote the true values.]{
\includegraphics[width=0.7\textwidth]{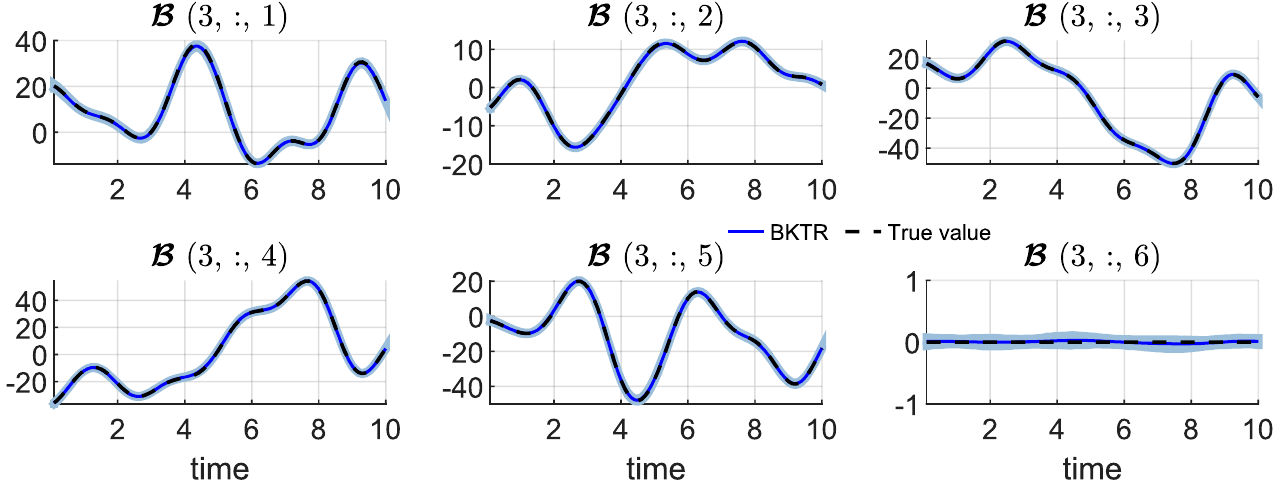}
}
\subfigure[Interpolated spatial surfaces of true value, estimated value (posterior mean), and absolute estimation error of the coefficients at time point $n=20$, i.e., $\boldsymbol{\mathcal{B}}(:,20,p=1,2,4,6)$. Black circles denote the positions of sampled locations.]{
\includegraphics[width=0.97\textwidth]{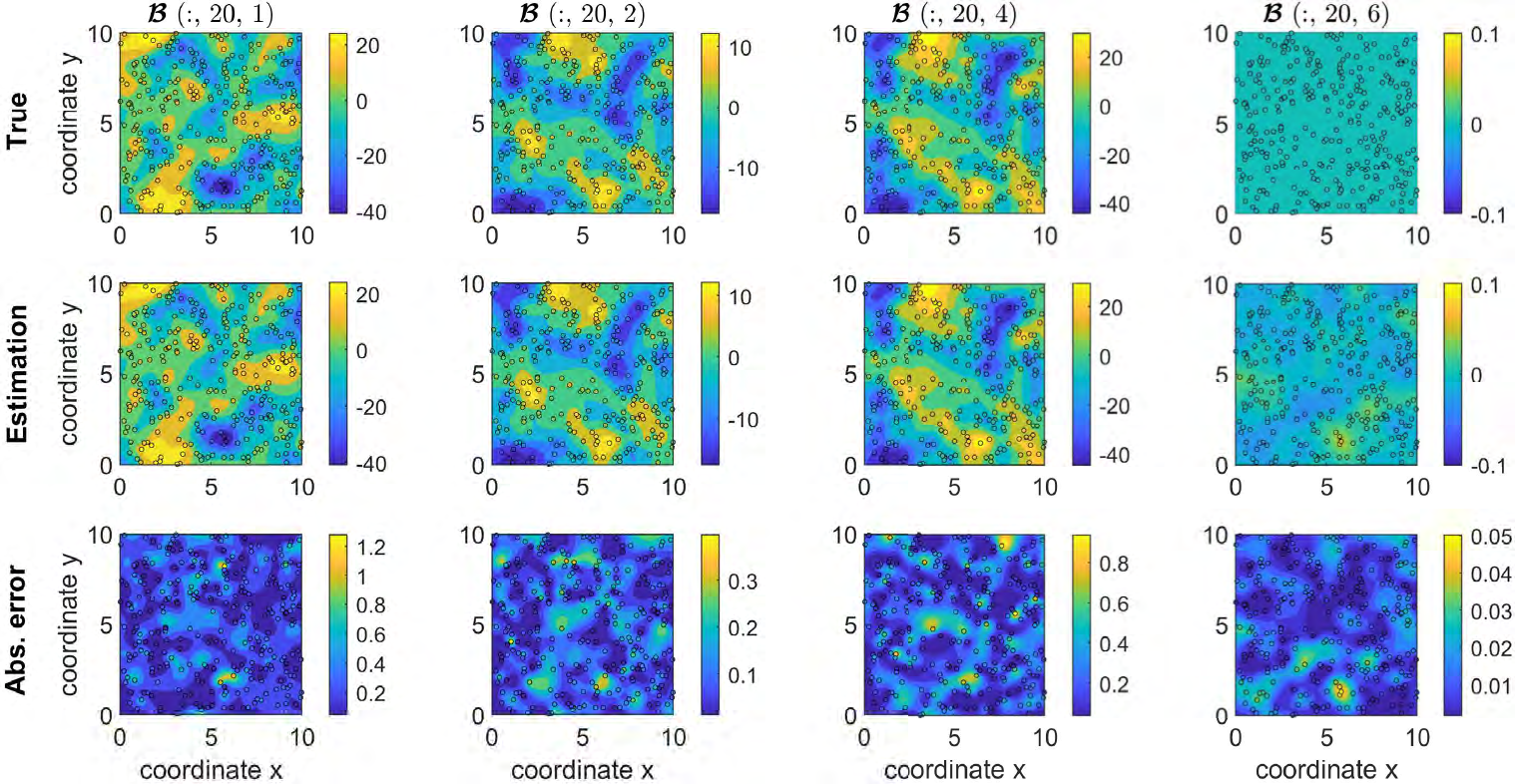}
}
\caption{BKTR ($R=10$) estimated coefficients for Simulation 1 (50\% of $\boldsymbol{Y}$ is observed).}
\label{fig:simu1_new}
\end{figure}

Table~\ref{tab:simu1} shows the performance metrics (mean$\pm$std calculated from the 40 replicated simulations) of BKTR in the case where the true rank is specified ($R=10$) and when 50\% of $\boldsymbol{Y}$ is partially observed. We can see that the CVG of 95\% CI of $\boldsymbol{\mathcal{B}}$ is around 95\% as expected. We also plot some estimation results of one chain/simulation in Figure~\ref{fig:simu1_new}, where panel (a) illustrates the temporal evolution of the estimated coefficients for each covariate at a given location, and panel (b) maps spatial surfaces of $\boldsymbol{\mathcal{B}}$ for each covariate at a given time point. These plots show that BKTR can effectively approximate the spatiotemporally varying relationships between the spatiotemporal data and the corresponding covariates.

\begin{figure}[!t]
\centering
\includegraphics[width=1\textwidth]{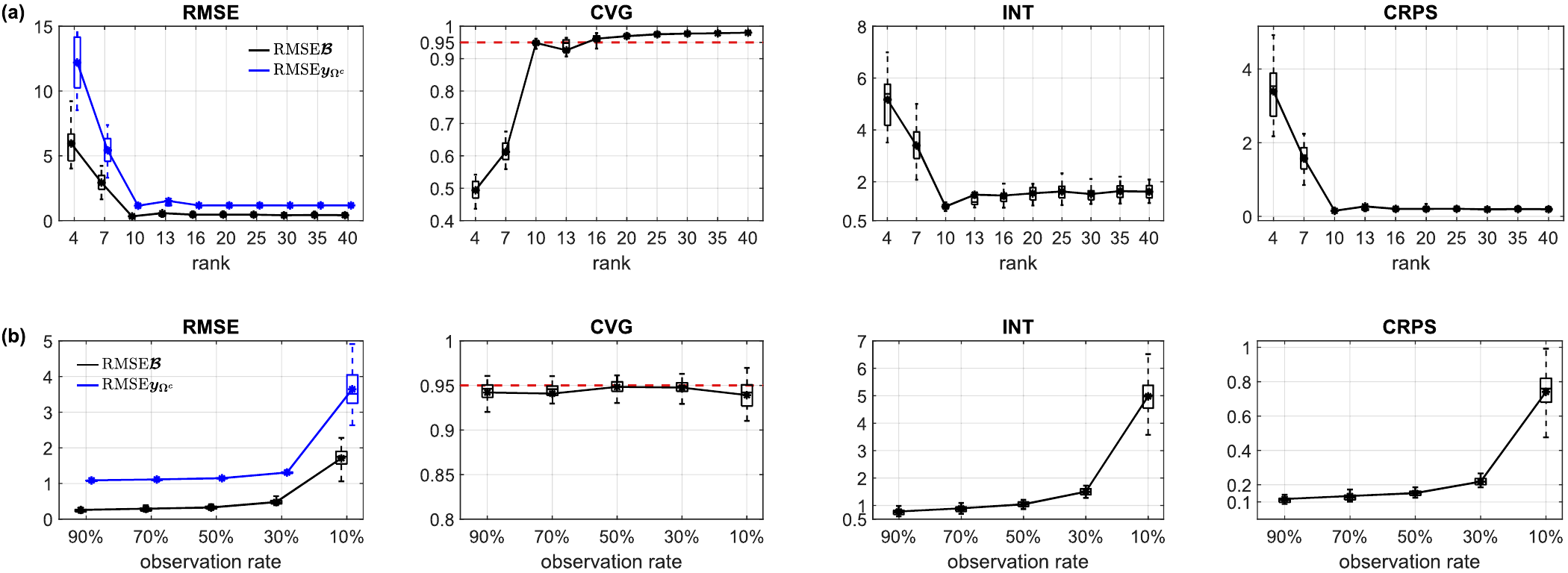}
\caption{Sensitivity test of BKTR for Simulation 1: (a) Effects of the rank $R$; (b) Effects of the observation rate $\frac{|\Omega|}{MN}$. For each case, the figure shows the boxplots and mean values of the corresponding metrics calculated from 40 replications.}
\label{fig:simu1_sensitivity}
\end{figure}

\begin{figure}[!t]
\centering
\subfigure[Coefficients ($\boldsymbol{\mathcal{B}}(3,:,6)$) estimated by BKTR with different rank settings (when 50\% of the data are observed).]{
\includegraphics[width=1\textwidth]{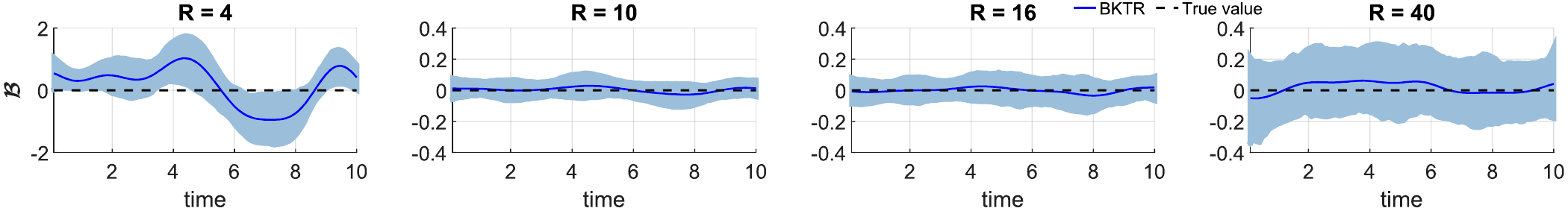}
}
\subfigure[Coefficients ($\boldsymbol{\mathcal{B}}(3,:,3)$) estimated by BKTR (with $R=10$) under different observation rates.]{
\includegraphics[width=1\textwidth]{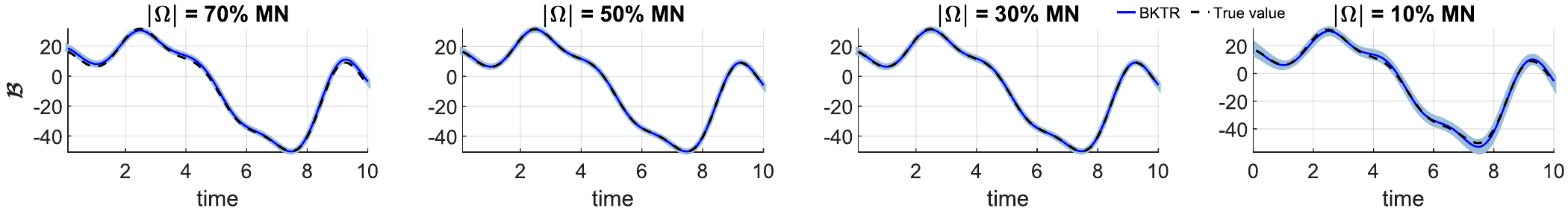}
}
\caption{Comparison of BKTR in different settings for Simulation 1. (a) and (b) plot the estimated $\boldsymbol{\mathcal{B}}$ (mean with 95\% CI) of one simulation at location \#3 ($m=3$) for the 6th and 3rd covariates (i.e., $\boldsymbol{\mathcal{B}}(3,:,6)\text{ and }\boldsymbol{\mathcal{B}}(3,:,3)$), respectively.}
\label{fig:simu1_new_compare}
\end{figure}

The performance of BKTR with different rank specifications and 50\% of the data points observed are compared in Figure~\ref{fig:simu1_sensitivity}(a). We see that when the rank is overspecified (larger than the true value, i.e., $R>10$), BKTR can still achieve reliable estimation accuracy for both coefficients and unobserved data, even when $R$ is much larger than 10, e.g., $R=40$. The CVG for the 95\% CI of $\boldsymbol{\mathcal{B}}$ is also maintained around 95\% when $R>10$, the corresponding INT and CRPS maintain a low value. This indicates that BKTR is able to provide accurate estimation for the coefficients along with high-quality uncertainty even when the model includes redundant parameters. In addition, Figure~\ref{fig:simu1_sensitivity}(b) illustrates the effect of the observation rate of $\boldsymbol{Y}$, where the results of BKTR (specifying $R=10$) with different proportions of observed data are compared. We observe that BKTR has a stable low estimation error when the rate of observed data decreases, and it continues to obtain valid CIs despite the increased lengths of CI for lower observation rates (i.e., the INT becomes larger). These results demonstrate the powerful applicability of the proposed Bayesian framework.
We further show the temporal estimations of coefficients at a given location obtained in different cases in Figure~\ref{fig:simu1_new_compare}. From Figure~\ref{fig:simu1_new_compare}(a), we see that a higher rank specification generates broader intervals, which is consistent with the INT results in Figure~\ref{fig:simu1_sensitivity}(a), i.e., the INT slightly increases with overspecified ranks. Panel (b) shows that BKTR can accurately estimate time-varying coefficients even when only 10\% of the response data is observed, demonstrating its high modeling capacity.

\subsection{Simulation 2: model comparison on a small data set}
\subsubsection{Simulation setting}
In this simulation, we generate a small data set in which the true $\boldsymbol{\mathcal{B}}$ is directly generated following a separable covariance matrix. Specifically, we simulate 30 locations $(M=30)$ in a $[0,10]\times[0,10]$ square and $N=30$ evenly distributed time points in $[0,10]$. We then generate a small synthetic data set ${\left\{\boldsymbol{Y}\in\mathbb{R}^{30\times30},\boldsymbol{\mathcal{X}}\in\mathbb{R}^{30\times30\times3}\right\} (P=3)}$ following Eq.~\eqref{Eq:STVC-vector} with:
\begin{equation} \notag
\begin{aligned}
    \bd{\cal{X}}(:,:,1)&=\operatorname{ones}(M,N),\
    \bd{\cal{X}}(:,:,2)=\boldsymbol{1}_{N}^{\top}\otimes\boldsymbol{x}_{s},\ \bd{\cal{X}}(:,:,3)=\boldsymbol{1}_{M}\otimes\boldsymbol{x}_{t}^{\top},
    \\    \operatorname{vec}\left(\boldsymbol{B}_{(3)}\right)&\sim\mathcal{N}\left(\boldsymbol{0},\boldsymbol{K}_{t}\otimes\boldsymbol{K}_{s}\otimes\boldsymbol{\Lambda}_{w}^{-1}\right),~\boldsymbol{\epsilon}\sim\mathcal{N}(\boldsymbol{0},\tau^{-1}\boldsymbol{I}_{MN}),
\end{aligned}
\end{equation}
where ${\boldsymbol{x}_{s}\sim\mathcal{N}(\bd{0},\boldsymbol{I}_{M})}$, ${\boldsymbol{x}_{t}\sim\mathcal{N}(\bd{0},\boldsymbol{I}_{N})}$, ${\boldsymbol{K}_{s}}$ and ${\boldsymbol{K}_{t}}$ are still computed from a Mat\'ern 3/2 kernel and a SE kernel, respectively, and ${\boldsymbol{\Lambda}_{w}\sim\mathcal{W}(\boldsymbol{I}_{P},P)}$. For model parameters, we set the kernel variance hyperparameters at $\sigma_s^2=\sigma_t^2=2$, and consider combinations of data noise variance $\tau^{-1}\in\{0.2,1,5\}$, and kernel length-scale hyperparameters $\phi=\gamma\in\{1,2,4\}$. We specify the CP rank ${R=10}$ for BKTR estimation and compare BKTR with the original STVC model \citep{gelfand2003spatial} and a pure low-rank Bayesian probabilistic tensor regression (BTR) model without imposing any spatiotemporal GP priors by setting $\boldsymbol{K}_{s}=\boldsymbol{I}_{M},\boldsymbol{K}_t=\boldsymbol{I}_N$ as the prior and using the same rank as BKTR.
For both BKTR and STVC, we assume that the kernel functions are known, i.e., $k_s$ is Mat\'ern 3/2 and $k_t$ is SE, and use MCMC to estimate the unknown kernel hyperparameters. As in the previous simulations, we add a normally distributed non-significant random covariate in $\boldsymbol{\mathcal{X}}$ when fitting the model. We replicate each simulation 25 times and run the MCMC sampling with $K_1=1000$ and $K_2=500$.

\subsubsection{Results}

\begin{table}[!t]
\footnotesize
    \centering
    \caption{{Performance comparison for Simulation 2 $\left(\text{MAE}_{\boldsymbol{\mathcal{B}}}/\text{RMSE}_{\boldsymbol{\mathcal{B}}}\right)$.}}
    \begin{tabular}{ll|ccc}
    \toprule
    Settings & & BTR & STVC & BKTR  \\
    \midrule
    \multirow{3}{*}{$\tau^{-1}=0.2$} & $\phi=\gamma=1$ & 1.00$\pm$0.29 / 1.57$\pm$0.49 & \textbf{0.64}$\pm$0.16 / \textbf{1.00}$\pm$0.29 & 0.81$\pm$0.19 / 1.32$\pm$0.32 \\
    & $\phi=\gamma=2$ & 0.94$\pm$0.35 / 1.54$\pm$0.62 & 0.63$\pm$0.15 / 1.09$\pm$0.23 & \textbf{0.61}$\pm$0.18 / \textbf{1.06}$\pm$0.34  \\
    & $\phi=\gamma=4$ & 0.74$\pm$0.29 / 1.24$\pm$0.54 & 0.53$\pm$0.11 / 0.88$\pm$0.19 & \textbf{0.40}$\pm$0.13 / \textbf{0.70}$\pm$0.28 \\
    \cmidrule(r){1-5}
    \multirow{3}{*}{$\tau^{-1}=1$} & $\phi=\gamma=1$ & 1.00$\pm$0.25 / 1.55$\pm$0.44 & 0.82$\pm$0.10 / 1.25$\pm$0.24 & \textbf{0.79}$\pm$0.18 / \textbf{1.23}$\pm$0.33 \\
    & $\phi=\gamma=2$ & 0.85$\pm$0.23 / 1.33$\pm$0.41 & 0.63$\pm$0.07 / 0.92$\pm$0.15 & \textbf{0.60}$\pm$0.12 / \textbf{0.90}$\pm$0.22 \\
    & $\phi=\gamma=4$ & 0.73$\pm$0.22 / 1.18$\pm$0.42 & 0.54$\pm$0.09 / 0.74$\pm$0.16 & \textbf{0.45}$\pm$0.09 / \textbf{0.71}$\pm$0.17 \\
    \cmidrule(r){1-5}
    \multirow{3}{*}{$\tau^{-1}=5$} & $\phi=\gamma=1$ & 1.10$\pm$0.16 / 1.66$\pm$0.27 & 1.09$\pm$0.12 / 1.60$\pm$0.19 & \textbf{0.89}$\pm$0.14 / \textbf{1.32}$\pm$0.23 \\
    & $\phi=\gamma=2$ & 1.01$\pm$0.16 / 1.52$\pm$0.30 & 0.92$\pm$0.12 / 1.31$\pm$0.18 & \textbf{0.75}$\pm$0.11 / \textbf{1.11}$\pm$0.20 \\
    & $\phi=\gamma=4$ & 0.90$\pm$0.15 / 1.37$\pm$0.29 & 0.78$\pm$0.08 / 1.08$\pm$0.14 & \textbf{0.61}$\pm$0.07 / \textbf{0.89}$\pm$0.14 \\
    \cmidrule(r){1-5}
    \multicolumn{2}{l|}{Average computing time} & $\approx$0.008 sec/iter. & $\approx$15.44 sec/iter. & $\approx$0.011 sec/iter. \\
    \bottomrule
    \multicolumn{4}{l}{{Best results are highlighted in bold fonts.}}
    \end{tabular}
    \label{tab:simu2}
\end{table}

\begin{figure}[!t]
\centering
\includegraphics[width=0.7\textwidth]{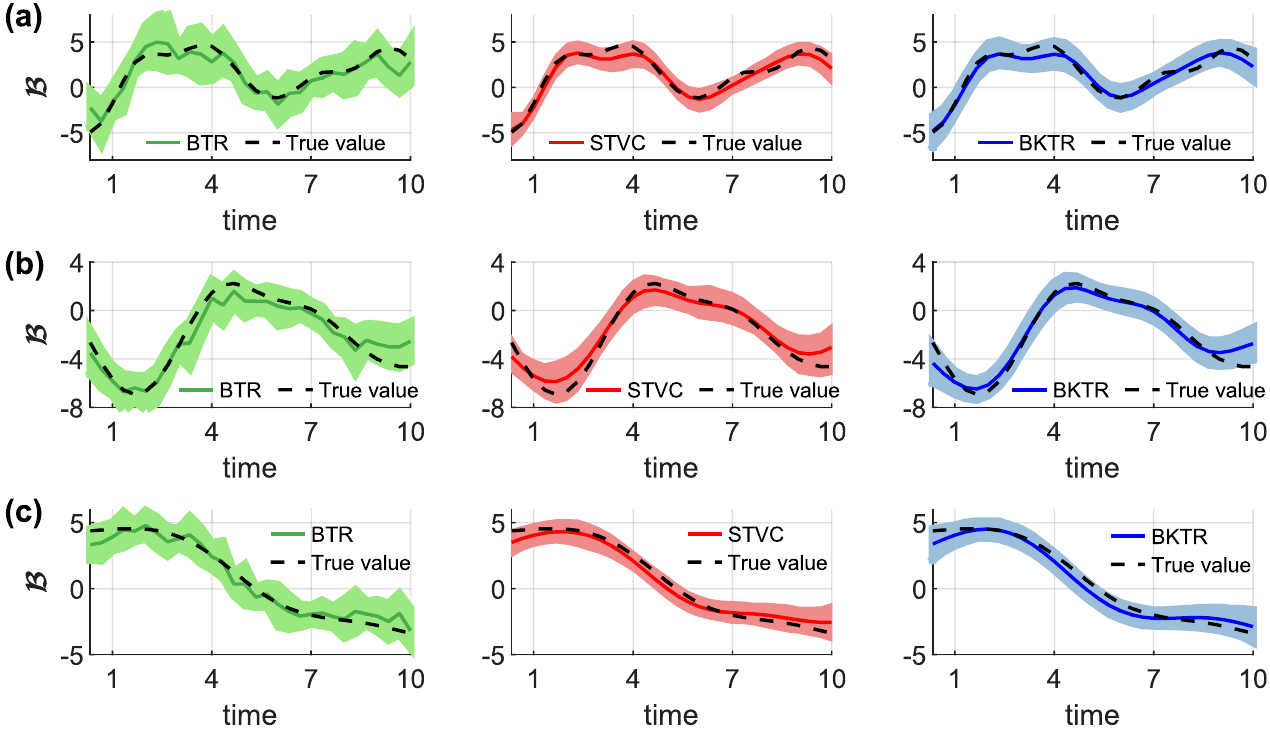}
\caption{Comparison of the estimated coefficients when (a) $\tau^{-1}=1,\phi=\gamma=1$, (b) $\tau^{-1}=1,\phi=\gamma=2$, (c) $\tau^{-1}=1,\phi=\gamma=4$. We show the approximated coefficients for the third covariate at location \#8, with solid curves representing the posterior mean and the shaded areas denoting 95\% CI.}
\label{fig:simu2_temporal}
\end{figure}

\begin{figure}[!t]
\centering
\includegraphics[width=1.05\textwidth]{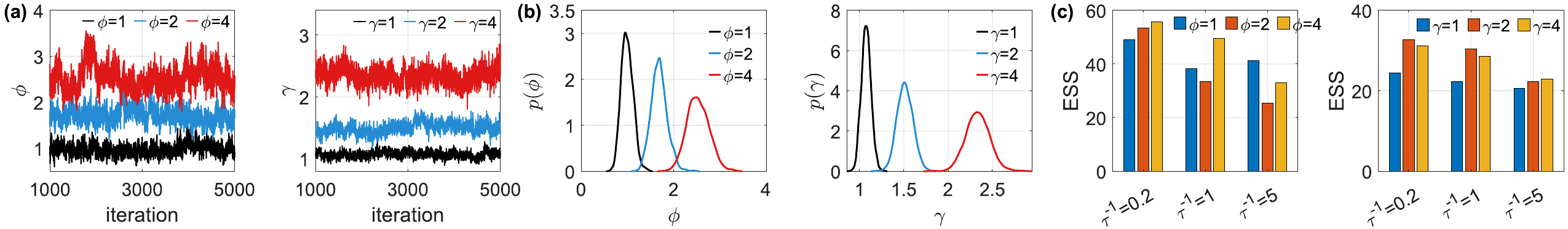}
\caption{BKTR estimated kernel hyperparameters for Simulation 2 when $\tau^{-1}=1$: (a) trace plots; (b) posterior densities. (c): ESS of kernel hyperparameters learned by BKTR for Simulation 2.}
\label{fig:simu2_HyperESS}
\end{figure}

Table~\ref{tab:simu2} presents the accuracy (mean$\pm$std) of the estimated coefficient tensor $\bd{\cal{B}}$ obtained from the 25 simulations. We also report the average computing time for each MCMC iteration. As we can see, BKTR with rank $R=10$ achieves similar accuracy to that of STVC, and even exhibits superior performance in the presence of increased data noise or enlarged kernel length-scales. This means that BKTR is more robust for noisy data and the low-rank assumption can fit the spatiotemporal relations better when they vary more smoothly/slowly. Additionally, BKTR also clearly outperforms BTR, which confirms the importance of introducing GPs to encode the spatial and temporal dependencies. These results suggest that the proposed kernelized CP decomposition can approximate the true coefficient tensor relatively well even if the true $\bd{\cal{B}}$ does not follow an exact low-rank specification. In terms of computing time, we see that BKTR is much faster than STVC. Due to the high cost of STVC, it becomes infeasible to analyze a data set of moderate size, e.g., $M,N=100$, and $P=10$.

Figure~\ref{fig:simu2_temporal} shows the estimation results of the three approaches for one chain/simulation when $\tau^{-1}=1, \phi=\gamma=\{1,2,4\}$. We plot an example for the third covariate at location \#8 ($m=8, p=3$) to show the temporal variation of the coefficients estimated by the three methods. As can be seen, BKTR and STVC generate similar results and most of the coefficients are contained in the 95\% CI for these two models. Although BTR can still capture the general trend, it fails to reproduce the local temporal dependency due to the absence of spatial/temporal priors. To further verify the mixing of the GP hyperparameters, we run the MCMC for BKTR for 5000 iterations and show the trace plots and probability distributions (after burn-in) for $\phi$ and $\gamma$ obtained when $\tau^{-1}=1, \phi=\gamma\in\{1,2,4\}$ in Figure~\ref{fig:simu2_HyperESS}(a) and (b), respectively. Figure~\ref{fig:simu2_HyperESS}(c) shows the effective sample size (ESS) of kernel hyperparameters obtained by BKTR in different settings, where in each case the mean ESS of 25 simulations are given. We can see that the Markov chains for the kernel hyperparameters mix fast and well, and a few hundred iterations should be sufficient for posterior inference.

\subsection{Simulation 3: STVC modeling on a moderate-sized data set}
\subsubsection{Simulation setting}
To evaluate the performance of BKTR in more realistic settings where the data size is large and the relationships are usually captured without low-rank structure, we further generate a relatively large synthetic data set of the same size as in Simulation 1, i.e., $M=300$ locations and $N=100$ time points. We generate the covariates $\boldsymbol{\mathcal{X}}\in\mathbb{R}^{300\times100\times5}~(P=5)$ according to Eq.~(\ref{Eq:X_generate}). The coefficients are simulated using $\operatorname{vec}\left(\boldsymbol{B}_{(3)}\right)\sim\mathcal{N}\left(\boldsymbol{0},\boldsymbol{K}_{t}\otimes\boldsymbol{K}_{s}\otimes\boldsymbol{\Lambda}_{w}^{-1}\right)$. The parameters $\left\{\boldsymbol{K}_{s},\boldsymbol{K}_{t},\boldsymbol{\Lambda}_{w}\right\}$ and $\boldsymbol{\epsilon}$ are specified as in Simulation 1, i.e., $\boldsymbol{K}_{s}$ and $\boldsymbol{K}_{t}$ are computed from a Mat\'ern 3/2 and a SE kernel, respectively, with hyperparamaters $\{\sigma_s^2=\sigma_t^2=2, \phi=\gamma=1\}$, $\boldsymbol{\Lambda}_{w}\sim\mathcal{W}(\boldsymbol{I}_{P},P)$, and $\tau^{-1}=1$. We randomly select 50\% of the generated data $\boldsymbol{Y}$ as observed responses. To assess the effect of the rank specification, we try different CP rank settings $R$ in $\{5,10,\ldots,60\}$ and compute MAE and RMSE for both $\boldsymbol{\mathcal{B}}$ and $\boldsymbol{y}_{\Omega^c}$. We replicate the experiment with each rank setting 15 times and run the MCMC sampling with $K_1=1000$ and $K_2=500$.

\subsubsection{Results}
\begin{figure}[!t]
\centering
\subfigure[Estimated coefficients (mean with 95\% CI) of 4 covariates at location $m=3$, i.e. $\boldsymbol{\mathcal{B}}(3,:,p=1,4,5,6)$.]{
\includegraphics[width=1\textwidth]{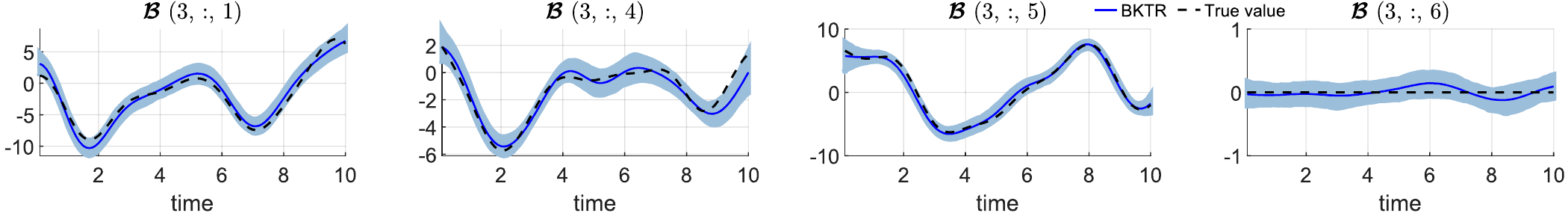}
}
\subfigure[Interpolated spatial surfaces of the estimated coefficients at time point $n=20$, i.e. $\boldsymbol{\mathcal{B}}(:,20,p=1,2,4,6)$, where black circles denote the positions of sampled locations.]{
\includegraphics[width=0.92\textwidth]{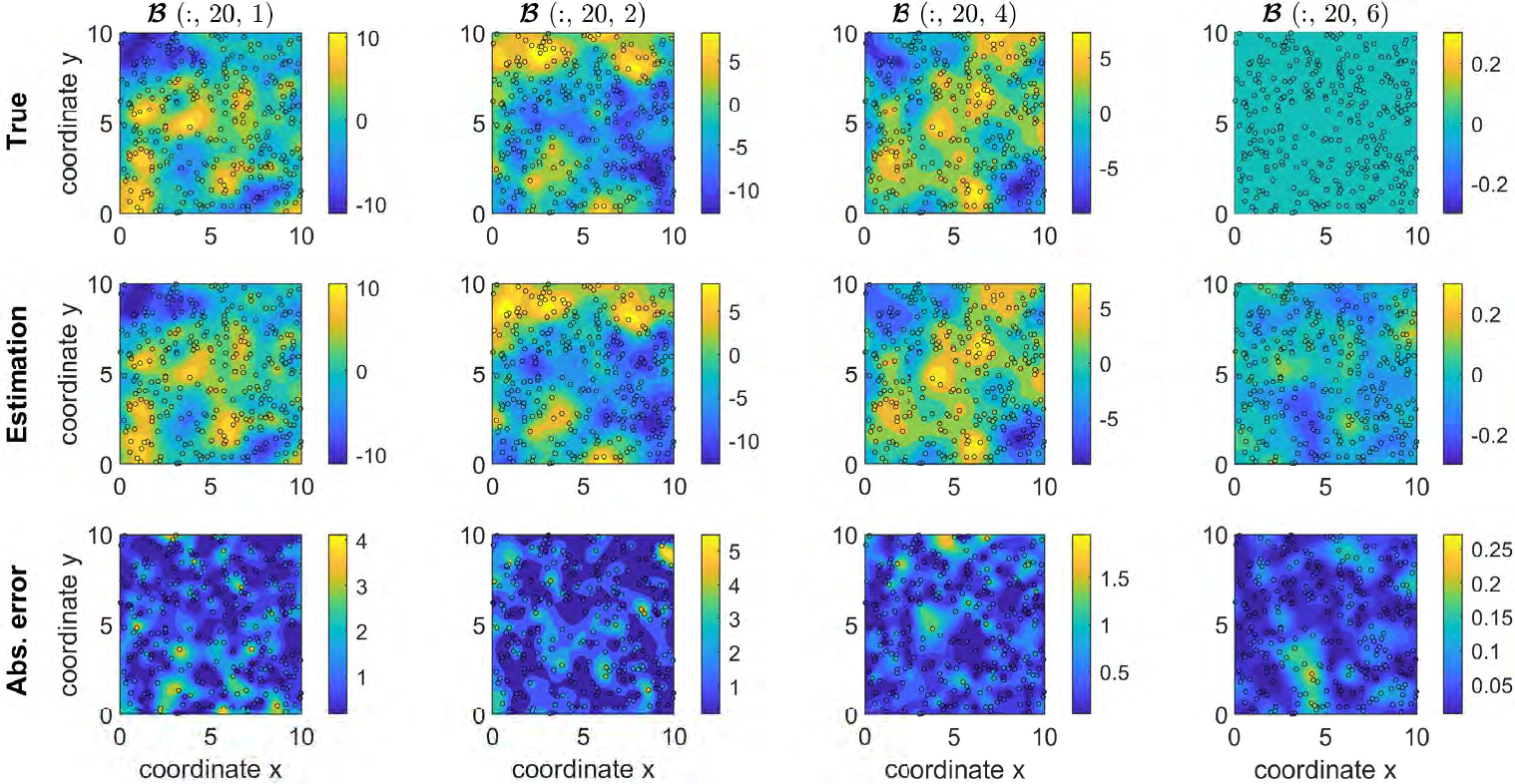}
}
\caption{BKTR ($R=40$) estimated coefficients for Simulation 3.}
\label{fig:simu3_new}
\end{figure}

\begin{figure}[!t]
\centering
\includegraphics[width=0.6\textwidth]{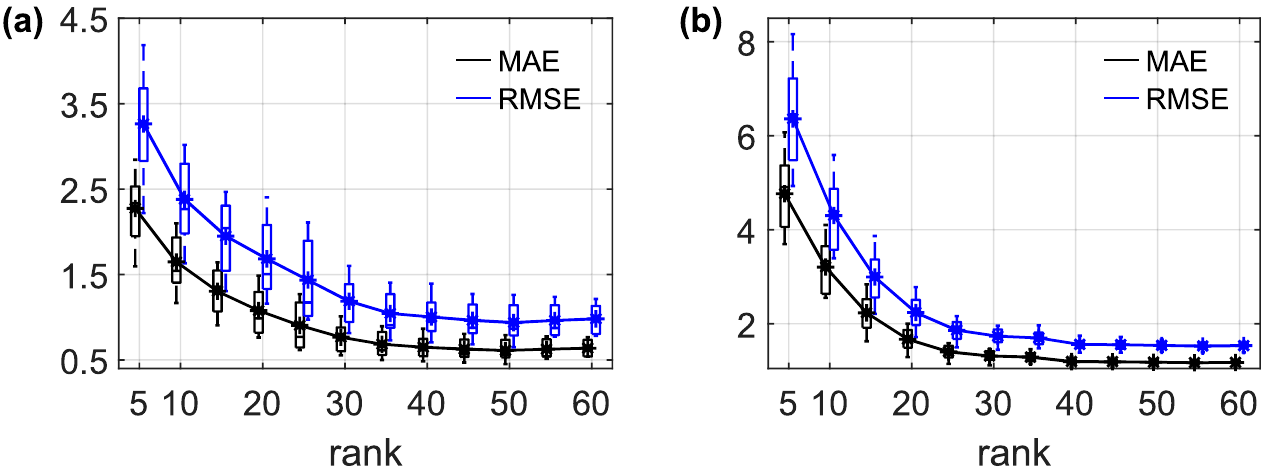}
\caption{The effect of rank for Simulation 3. The figure shows the boxplot and mean of the estimation error on (a) $\boldsymbol{\mathcal{B}}$ and (b) $\boldsymbol{y}_{\Omega^c}$ with respect to the tensor rank.}
\label{fig:simu3_rank}
\end{figure}

Figure~\ref{fig:simu3_new} shows the temporal behaviors and spatial patterns of the estimated coefficients of four covariates in one simulation when $R=40$.
As one can see, BKTR can effectively reproduce the true values with acceptable errors. Figure~\ref{fig:simu3_rank} shows the effect of rank. We see that choosing a larger rank $R$ gives better accuracy in terms of both $\text{MAE}_{\boldsymbol{\mathcal{B}}}/\text{RMSE}_{\boldsymbol{\mathcal{B}}}$ and $\text{MAE}_{\boldsymbol{y}_{\Omega^c}}/\text{RMSE}_{\bd{{y}}_{\Omega^c}}$. However, the accuracy gain becomes marginal when $R>30$. This demonstrates that the proposed Bayesian treatment offers a flexible solution in terms of model estimation.

\section{Bike-sharing Demand Modeling}
\label{sec:application}
\subsection{Data description}
In this section, we perform local spatiotemporal regression on a large-scale bike-sharing trip data set collected from BIXI\footnote{\url{https://bixi.com}}---a docked bike-sharing service in Montreal, Canada. BIXI operates 615 stations across the city and the yearly service time window is from April to November. We collect the number of daily departures for each station over $N=196$ days (from April 15 to October 27, 2019). We discard the stations with an average daily demand of less than 5, and only consider the remaining $M=587$ stations. The matrix $\bd{Y}$ contains $13.0\%$ unobserved/corrupted values in the raw data, and we only keep the rest as the observed set $\Omega$. {Given that the collected daily departures can have different variances across the spatial locations, we take the logarithm of the data matrix $\boldsymbol{Y}$ to make the variances more consistent and obtain a nearly Gaussian distributed data. We approximate this transformed dataset using the assumption of Gaussian likelihood with a linear covariance and a homoscedastic residual process.

We follow the analysis in \citet{faghih2014land} and \citet{wang2021modeling}, and consider two types of spatial covariates: (a) features related to cycling infrastructure, and (b) land-use and built environment features. Particularly, similar to the operations in \citet{wang2021modeling}, the spatial covariates are collected using the intersections of 250-meter circle buffers and Thiessen polygons for obtaining predictors from non-overlapping areas. For temporal covariates, we mainly consider weather conditions and holidays. Table~\ref{tab:variables} lists the 13 spatial covariates $(\boldsymbol{x}_{s}^{p}\!\in\!\mathbb{R}^{M})$ and 5 temporal covariates $(\boldsymbol{x}_{t}^{p}\!\in\!\mathbb{R}^{N})$ used in this analysis. The final number of covariates is $P\!=\!13+5+1\!=\!19$, including the intercept. In constructing the covariate tensor $\boldsymbol{\mathcal{X}}$, we follow the same approach as in the simulation experiments. Specifically, we set $\bd{\cal{X}}(:,:,1)\!=\!\operatorname{ones}(M,N)$, and fill the rest of the tensor slices with $\boldsymbol{1}_{N}^{T}\otimes\boldsymbol{x}_{s}^{p}~(p=2:14)$ for the spatial covariates and $\boldsymbol{1}_{M}\otimes(\boldsymbol{x}_{t}^{p})^{\top}~(p=15:19)$ for the temporal covariates. Given the difference in magnitudes, we normalize all covariates to $[0,1]$ using a min-max normalization. Since we use a zero-mean GP to parameterize $\bd{\beta}$, we normalize departure trips by removing the global effect of all covariates through linear regression and consider $\bd{y}$ to be the unexplained part. The coefficient tensor $\bd{\cal{B}}$ contains more than $2\times 10^6$ entries, preventing us from using STVC directly.

\begin{table}[!ht]
\small
    \centering
    \caption{Description summary of independent variables.}
    \begin{tabular}{ll|p{10cm}}
    \toprule
    \multirow{2}{*}{spatial} & (a) & length of cycle paths; length of major roads; length of minor roads; station capacity; \\
    \cmidrule(r){2-3}
    & (b) & numbers of metro stations, bus stations, and bus routes; numbers of restaurants, universities, other business \& enterprises; park area; walkscore; population; \\
    \midrule
    \multicolumn{2}{l|}{temporal}  & daily relative humidity; daily maximum temperature;  daily mean temperature; total precipitation; dummy variables for holidays. \\
    \bottomrule
    \end{tabular}
    \label{tab:variables}
\end{table}

\subsection{Experimental setup}
For spatial factors, we use a Mat\'ern 3/2 kernel as a universal prior, i.e., $k_s(\bd{s}_{m},\bd{s}_{m'})=\left(1+\frac{\sqrt{3}d}{\phi}\right)\exp{\left(-\frac{\sqrt{3}d}{\phi}\right)}$, where $d$ is the Euclidean distance between locations $\boldsymbol{s}_{m}$ and $\boldsymbol{s}_{m'}$, and $\phi$ is the spatial length-scale. The Mat\'ern class kernel is commonly used as a prior kernel function in spatial modeling. For temporal factors, we use a locally periodic correlation matrix by taking the product of a periodic kernel and a SE kernel, i.e., $k_t(t_{n},t_{n'})=\exp{\left(-\frac{2\sin^2{(\pi(t_n-t_{n'})/T)}}{\gamma_{1}^{2}}-\frac{(t_n-t_{n'})^2}{2\gamma_{2}^{2}}\right)}$, where $\gamma_1$ and $\gamma_2$ denote the length-scale and decay-time for the periodic component, respectively, and we fix the period as $T=7$ days. This specification suggests a weekly temporal pattern that can change over time and allows us to characterize both the daily variation and the weekly trend of the coefficients. We set the CP rank $R=20$, and run MCMC with  $K_1=1000$ and $K_2=500$ iterations.

\subsection{Results}

\begin{figure}[!t]
\centering
\subfigure[Temporal plots of coefficients for 3 spatial covariates at location \#24 $(m=24)$.]{
\includegraphics[width=1\textwidth]{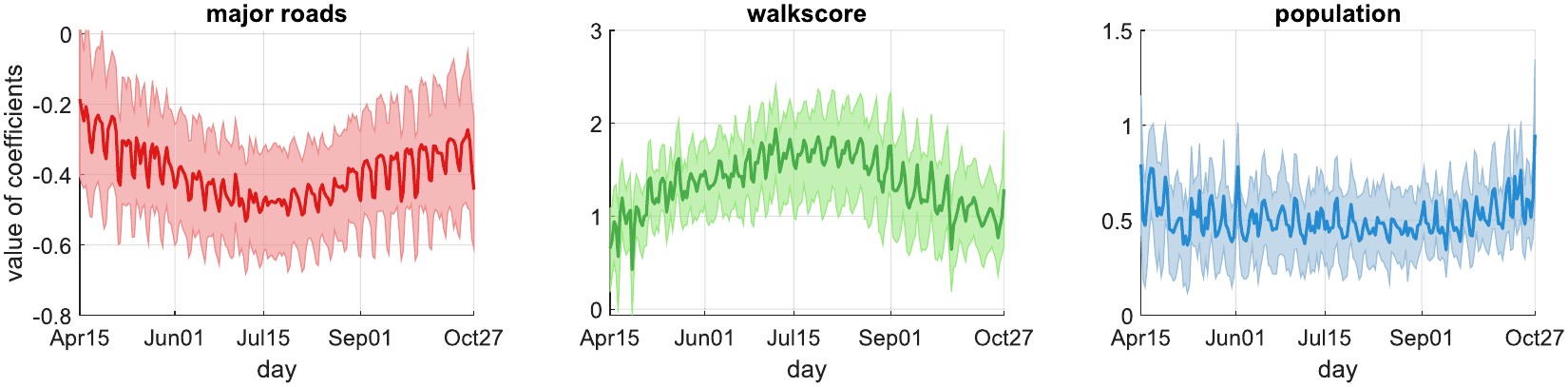}
}
\subfigure[Temporal plots of coefficients for 3 temporal covariates at location \#7 $(m=7)$.]{
\includegraphics[width=1\textwidth]{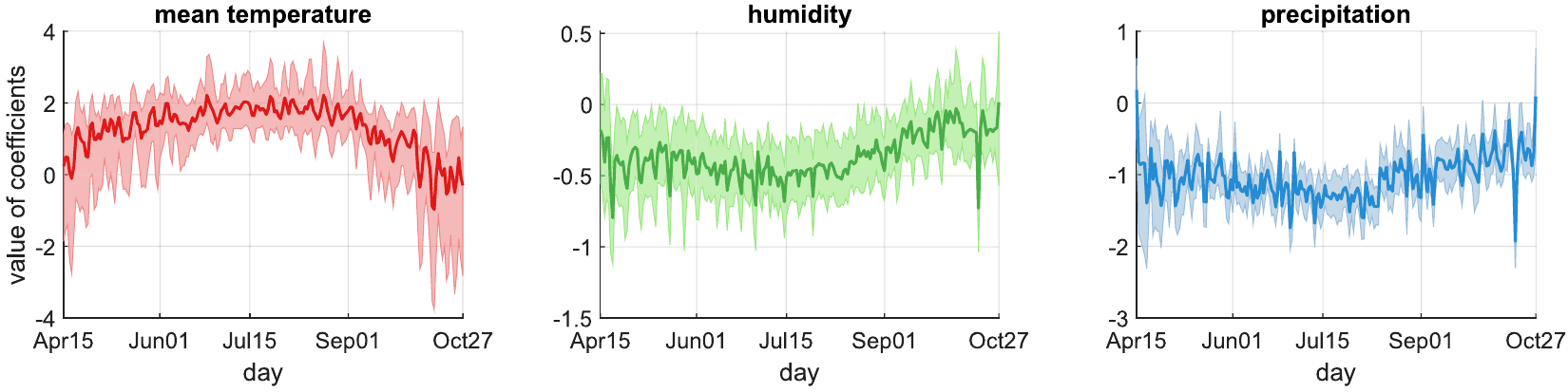}
}
\caption{Temporal illustration of BKTR estimated coefficients for log-transformed BIXI demand data. The plots show the posterior mean with 95\% CI.}
\label{Fig:logBIXICoe_TemPlots}
\end{figure}

\begin{figure}[!t]
\centering
\subfigure[Spatial maps of coefficients mean.]{
\includegraphics[width=0.47\textwidth]{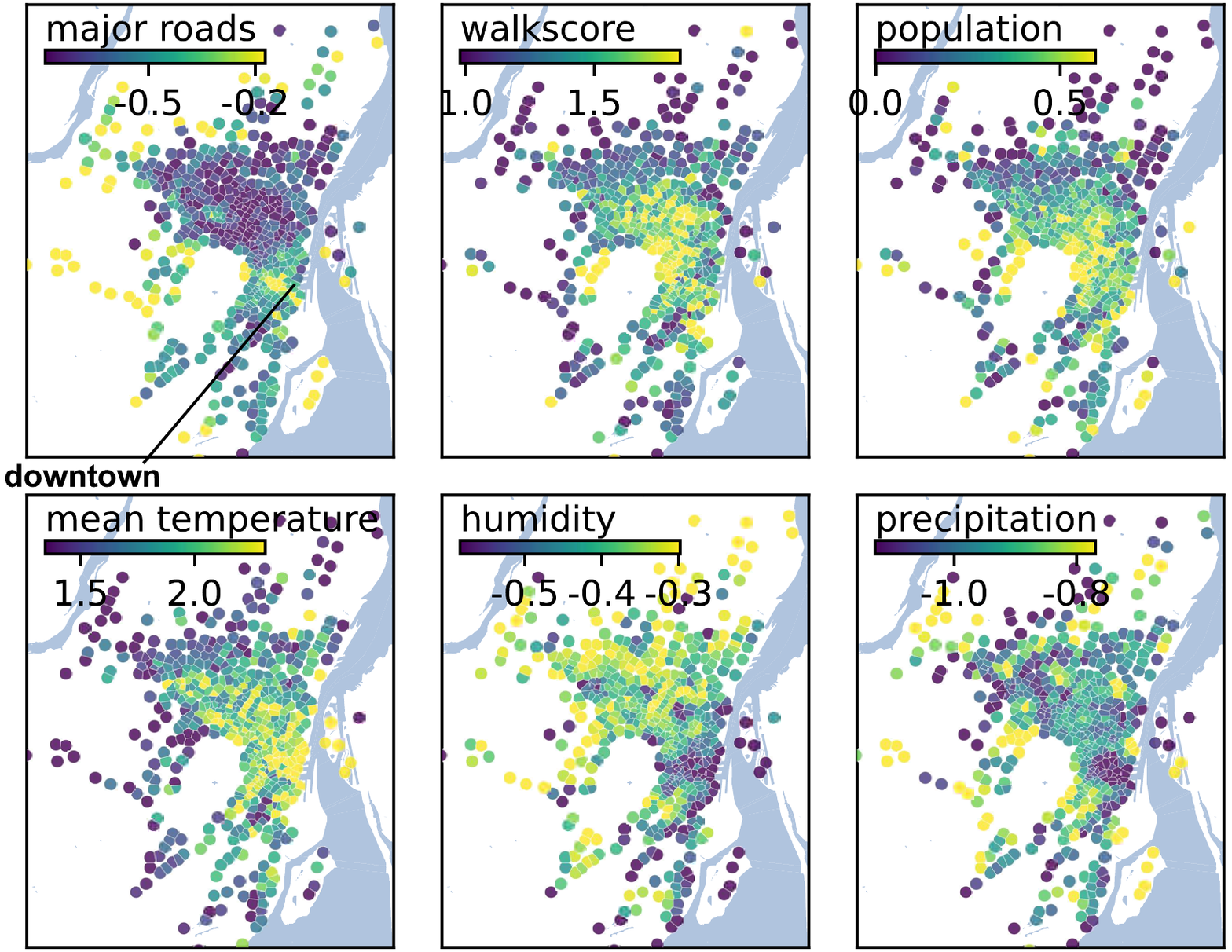}
}
\subfigure[Spatial maps of coefficients std.]{
\includegraphics[width=0.47\textwidth]{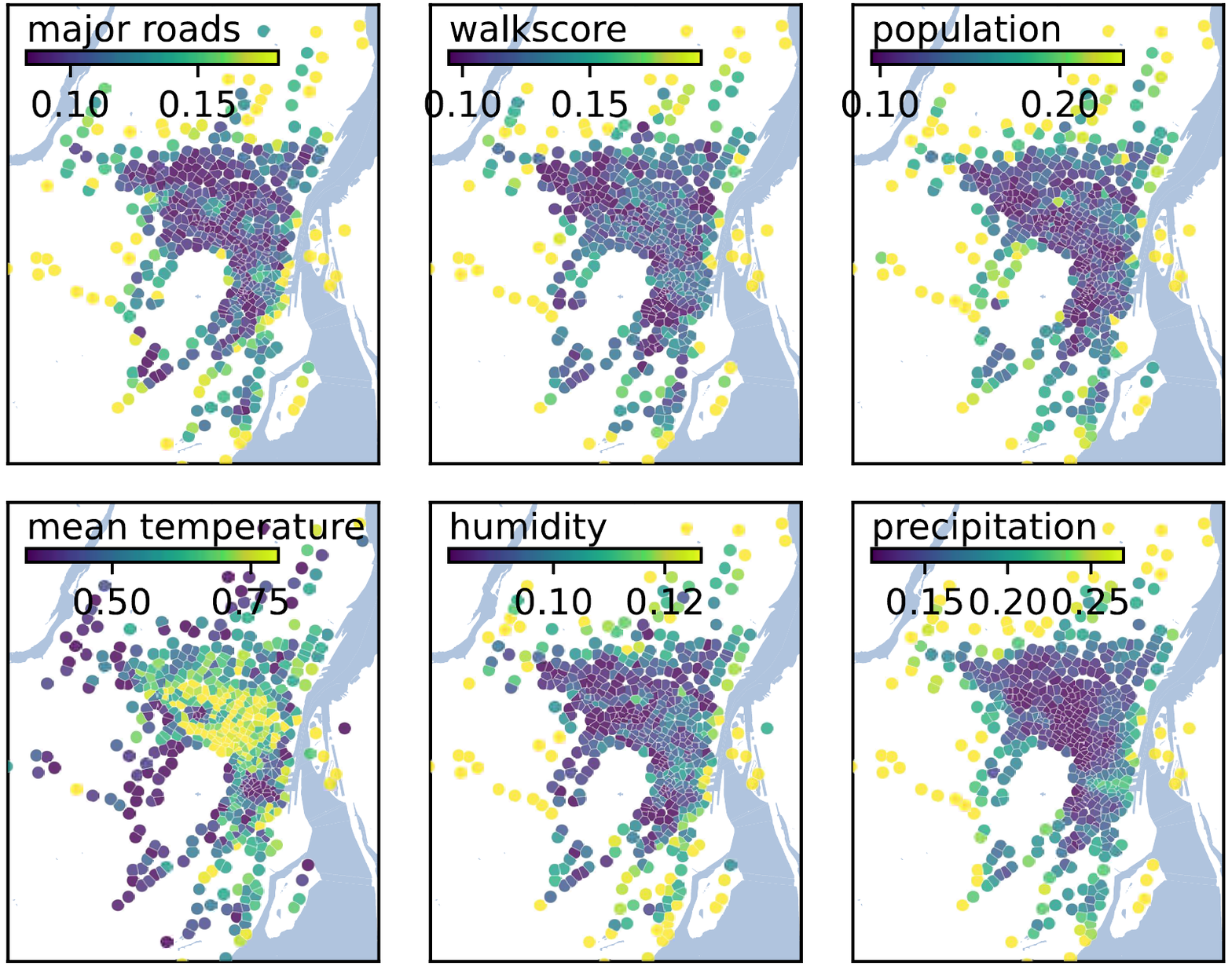}
}
\caption{Spatial maps of BKTR estimated coefficients for log-transformed BIXI demand: (a) posterior mean for 6 covariates on August 1, 2019 $(n=109)$; (b): the corresponding posterior standard deviation (std.).}
\label{Fig:logBIXICoe_SpaMaps}
\end{figure}

Figure~\ref{Fig:logBIXICoe_TemPlots} and~\ref{Fig:logBIXICoe_SpaMaps} demonstrate several examples of estimated coefficients, showing temporal plots of $\bd{\cal{B}}(\bd{s}_{m},:,:)$ and spatial maps of $\bd{\cal{B}}(:,t_{n},:)$, respectively. As we can see in Figure~\ref{Fig:logBIXICoe_TemPlots}, the temporal variations of the coefficients for both spatial and temporal covariates are interpretable. The coefficients (along with CI describing the uncertainty) allow us to identify the most important factors for each station at each time point. For example, we observe a similar variation over a week and a general long-term trend from April to October. Furthermore, the magnitude of the coefficients can be larger during summer (July / August) compared to the beginning and end of the operation period (e.g., for major roads, walkscore, and mean temperature), which is as expected as the outdoor temperature drops. Overall, for spatial variables, the positive correlation of walkability and the negative impact caused by the length of major roads indicate that bicycle demands tend to be higher in densely populated neighborhoods. For the temporal covariates, the precipitation and humidity variables both relate to negative coefficients, implying that people are less likely to ride bicycles in rainy/humid periods. The spatial distributions of the coefficients in Figure~\ref{Fig:logBIXICoe_SpaMaps} also demonstrate consistent estimations, where the effects of covariates tend to be more obvious in the downtown area, involving coefficients with larger absolute values. Again, one can further explore the credible intervals to evaluate the significance of each covariate at a given location. In addition, the estimated variance $\tau^{-1}$ for $\boldsymbol{\epsilon}$ gives a posterior mean of $6.19\times 10^{-2}$ with a 95\% CI of $[6.15, 6.28]\times 10^{-2}$. This variance is much smaller compared with the variability of the data, validating a good performance of data fitting using the proposed BKTR model. In summary, BKTR produces interpretable results for understanding the effects of different spatial and temporal covariates on bike-sharing demand. These findings can help planners evaluate the performance of bike-sharing systems and update/design them.

\subsection{Spatial interpolation}
Due to the introduction of GP, BKTR is able to perform spatial prediction, i.e., kriging, for unknown locations, and such results can be used to select sites for new stations. To validate the spatial prediction/interpolation capacity, we randomly select 30\% locations of the BIXI data set to be missing, and compute the prediction accuracy for these missing/testing data ($\boldsymbol{y}_{\Omega^c}$). Given that STVC is not applicable for a data set of such size, we compare BKTR (rank $R=20$) with SVC (spatially varying coefficient processes model) \citep{gelfand2003spatial} which omits the modeling in time dimension. A linear regression model with static $\boldsymbol{\beta}$ is also considered as the baseline in order to demonstrate the role of incorporating spatiotemporally varying coefficients in such prediction tasks. We also test these methods in the case where both 30\% locations and 30\% time periods are missing. The prediction errors are given in Table~\ref{tab:predictionBIXI}, where we show the MAE, RMSE and $R^2$ on $\boldsymbol{y}_{\Omega^c}$. $R^2$ is used to illustrate the fitting performance of the model, and the definition is also provided in Appendix~\ref{appC}. It is obvious that BKTR offers better performance with lower estimation errors in both missing scenarios, indicating its effectiveness in spatial prediction for real-world spatiotemporal data sets that often contain complicated missing/corruption patterns.

\begin{table}[!t]
\footnotesize
\centering
\caption{\small{Prediction performance for BIXI bike-sharing demand $\left(\operatorname{MAE}_{\boldsymbol{y}_{\Omega^c}}/\text{RMSE}_{\boldsymbol{y}_{\Omega^c}}/R^2\right)$.}}
\begin{tabular}{l|ccc}
\toprule
Scenarios & linear regression & SVC & BKTR ($R=20$) \\
\midrule
30\% location missing &  0.51/ 0.65/ 0.07  & 0.57/ 0.75/ 0.10  & \textbf{0.47}/ \textbf{0.63}/ \textbf{0.26}   \\
30\% location \& time points missing & 0.52/ 0.66/ 0.04  & 0.44/ 0.62/ 0.14  & \textbf{0.40}/ \textbf{0.57}/ \textbf{0.34} \\
\bottomrule
\multicolumn{3}{l}{{Best results are highlighted in bold fonts.}}
\end{tabular}
\label{tab:predictionBIXI}
\end{table}

\section{Discussion}
\label{sec: discuss}
\subsection{Identifiability and convergence of the model}

In the regression problem studied in this paper, the kernel/covariance hyperparameters, $\text{i.e., }\{\phi,\gamma\}$, interpret/imply the correlation structure of the data, and the coefficients, i.e., $\boldsymbol{\mathcal{B}}$, reveal the spatiotemporal processes of the underlying associations. Thus, we consider the convergence of kernel hyperparameters and the identifiability of the coefficients to be crucial. Note that the identifiability of latent factors decomposed from $\boldsymbol{\mathcal{B}}$ are not that important. For instance, our model is invariant in applying the same column permutation to the three factor matrices. From the results in simulation experiments (see Figure~\ref{fig:simu2_HyperESS}), it is clear that the Markov chains for kernel hyperparameters converge fast and well when using the proposed BKTR model.

\subsection{The superiority of the BKTR framework}
As we can see from the test of different rank settings and observation rates in the simulation experiments (see Figures~\ref{fig:simu1_sensitivity} and \ref{fig:simu3_rank}), BKTR is able to provide high estimation accuracy and valid CIs even with a much larger or over-specified rank, and also effectively estimates the coefficients and the unobserved output values when only 10\% of the data is observed. This indicates the advantage of the proposed Bayesian low-rank framework. Since we introduce a fully Bayesian sampling treatment for the kernelized low-rank tensor model which is free from parameter tuning, BKTR can estimate the model parameters and hyperparameters even when only a small number of observations are available. Thus, the model can consistently offer reliable estimation results, implying its effectiveness and usability for real-world complex spatiotemporal data analysis.
Another benefit of the proposed framework is the highly improved computing efficiency. As we mentioned in Section~\ref{sec_scalability}, the computational cost of BKTR is substantially reduced compared to the STVC model. According to the experiments conducted, BKTR is capable of dealing with regression problems containing up to millions of coefficients.

\section{Conclusion}
\label{sec: conclu}
This paper introduces an effective solution for large-scale local spatiotemporal regression analysis. We propose parameterizing the model coefficients using low-rank CP decomposition, which greatly reduces the number of parameters from $M\times N\times P$ to $R(M+N+P)$. Contrary to previous studies on tensor regression, the proposed model BKTR goes beyond the low-rank assumption by integrating GP priors to characterize the strong local spatial and temporal dependencies. The framework also learns a low-rank multi-linear kernel which is expressive and able to provide insights for nonstationary and complicated processes. Our numerical experiments on both synthetic data and real-world data suggest that BKTR can reproduce the local spatiotemporal processes efficiently and reliably.

There are several directions for future research. In the current model, the CP rank $R$ needs to be specified in advance. One can make the model more flexible and adaptive by introducing a reasonably large core tensor with a multiplicative gamma process prior such as in \citet{rai2014scalable}. In terms of GP priors, BKTR is flexible and can accommodate different kernels (w.r.t. function form and hyperparameter) for different factors such as in \citet{luttinen2009variational}. The combination of different kernels can also produce richer spatiotemporal dynamics and multiscale properties. In terms of computation, one can further reduce the cost in GP learning (e.g., $\mathcal{O}(M^3)$ for a spatial kernel) by further introducing sparse approximation techniques such as inducing points and predictive processes \citep{quinonero2005unifying,banerjee2008gaussian}. {Lastly, we can extend the regression model to handle binary- and count-valued data by using proper link functions. In this case, the challenge is that we no longer have analytical posteriors for latent factors; a potential solution is to use elliptical slice sampling \citep{murray2010elliptical} to sample the latent variables.}

\begin{acks}[Acknowledgments]
This research is supported by the Natural Sciences and Engineering Research Council of Canada (NSERC). Mengying Lei would like to thank the Institute for Data Valorization (IVADO) for providing a scholarship to support this study.
\end{acks}

\begin{supplement}
\addcontentsline{toc}{section}{Appendices}
\renewcommand{\thesubsection}{\Alph{subsection}}
\subsection{Calculation of the marginal likelihood}\label{appA}
We use the Woodbury matrix identity ${\footnotesize{(\boldsymbol{A}+\boldsymbol{C}\boldsymbol{B}\boldsymbol{C}^{\top})^{-1}=\boldsymbol{A}^{-1}-\boldsymbol{A}^{-1}\boldsymbol{C}\left(\boldsymbol{B}^{-1}+\boldsymbol{C}^{\top}\boldsymbol{A}^{-1}\boldsymbol{C}\right)^{-1}}}\\{\footnotesize{\boldsymbol{C}^{\top}\boldsymbol{A}^{-1}}}$ to compute $\boldsymbol{K}_{\left.\boldsymbol{y}\right|\phi}^{-1}$, and further calculate $\boldsymbol{y}_{\Omega}^{\top}\boldsymbol{K}_{\left.\boldsymbol{y}\right|\phi}^{-1}\boldsymbol{y}_{\Omega}$ in Eq.~\eqref{Eq:Marginal} as follows:
\begin{equation}\notag
\begin{aligned}
\boldsymbol{y}_{\Omega}^{\top}\boldsymbol{K}_{\left.\boldsymbol{y}\right|\phi}^{-1}\boldsymbol{y}_{\Omega}&=\boldsymbol{y}_{\Omega}^{\top}\left(\boldsymbol{H}_{U}\boldsymbol{K}_{U}\boldsymbol{H}_{U}^{\top}+\tau^{-1}\boldsymbol{I}_{|\Omega|}\right)^{-1}\boldsymbol{y}_{\Omega} \\
&=\boldsymbol{y}_{\Omega}^{\top}\left(\tau\boldsymbol{I}_{|\Omega|}-\tau^2\boldsymbol{H}_{U}\left(\boldsymbol{K}_{U}^{-1}+\tau\boldsymbol{H}_{U}^{\top}\boldsymbol{H}_{U}\right)^{-1}\boldsymbol{H}_{U}^{\top}\right)\boldsymbol{y}_{\Omega} \\
&=\tau\boldsymbol{y}_{\Omega}^{\top}\boldsymbol{y}_{\Omega}-\tau^2\boldsymbol{y}_{\Omega}^{\top}\boldsymbol{H}_{U}\left(\boldsymbol{I}_{R}\otimes\boldsymbol{K}_{s}^{-1}+\tau\boldsymbol{H}_{U}^{\top}\boldsymbol{H}_{U}\right)^{-1}\boldsymbol{H}_{U}^{\top}\boldsymbol{y}_{\Omega},
\end{aligned}
\end{equation}
where $\boldsymbol{H}_{U}=\boldsymbol{O}\tilde{\bd{X}}_{U}\left((\boldsymbol{W}\odot\boldsymbol{V})\otimes\boldsymbol{I}_{M}\right)$ with $\tilde{\boldsymbol{X}}_{U}=\left(\boldsymbol{X}_{(3)}\odot\boldsymbol{I}_{MN}\right)^{\top}$. Based on the matrix determinant lemma $\left|\boldsymbol{A}+\boldsymbol{C}\boldsymbol{B}\boldsymbol{C}^{\top}\right|=\left|\boldsymbol{B}^{-1}+\boldsymbol{C}^{\top}\boldsymbol{A}^{-1}\boldsymbol{C}\right|\det{(\boldsymbol{B})}\det{(\boldsymbol{A})}$, the determinant in Eq.~\eqref{Eq:Marginal}, i.e., $\log{\left|\boldsymbol{K}_{\left.\boldsymbol{y}\right|\phi}\right|}$, is computed as below:
\begin{equation} \notag
\begin{aligned}
\log{\left|\boldsymbol{K}_{\left.\boldsymbol{y}\right|\phi}\right|}&=\log{\left|\boldsymbol{H}_{U}\boldsymbol{K}_{U}\boldsymbol{H}_{U}^{\top}+\tau^{-1}\boldsymbol{I}_{|\Omega|}\right|} \\
&=\log{\left|\boldsymbol{I}_{R}\otimes\boldsymbol{K}_{s}^{-1}+\tau\boldsymbol{H}_{U}^{\top}\boldsymbol{H}_{U}\right|}+R\log{\left|\boldsymbol{K}_{s}\right|}-|\Omega|\log{\tau}.
\end{aligned}
\end{equation}
Therefore, the log marginal posterior of $\phi$ is updated with:
\begin{equation} \notag
\begin{aligned}
\log{p\left(\phi\given\boldsymbol{y}_{\Omega},\boldsymbol{V},\boldsymbol{W},\tau,\boldsymbol{\mathcal{X}}\right)}\propto&\log{p(\phi)}-\frac{1}{2}\boldsymbol{y}_{\Omega}^{\top}\boldsymbol{K}_{\left.\boldsymbol{y}\right|\phi}^{-1}\boldsymbol{y}_{\Omega}-\frac{1}{2}\log{\left|\boldsymbol{K}_{\left.\boldsymbol{y}\right|\phi}\right|} \\
\propto&\log{p(\phi)}+\frac{1}{2}\tau^2\boldsymbol{y}_{\Omega}^{\top}\boldsymbol{H}_{U}\left(\boldsymbol{I}_{R}\otimes\boldsymbol{K}_{s}^{-1}+\tau\boldsymbol{H}_{U}^{\top}\boldsymbol{H}_{U}\right)^{-1}\boldsymbol{H}_{U}^{\top}\boldsymbol{y}_{\Omega} \\
&-\frac{1}{2}\log{\left|\boldsymbol{I}_{R}\otimes\boldsymbol{K}_{s}^{-1}+\tau\boldsymbol{H}_{U}^{\top}\boldsymbol{H}_{U}\right|}-\frac{R}{2}\log{\left|\boldsymbol{K}_{s}\right|}.
\end{aligned}
\end{equation}

\subsection{Sampling algorithm for kernel hyperparameters}\label{appB}
The detailed sampling process for kernel hyperparmameters is provided in Algorithm~\ref{alg:HyperSample}.

\begin{algorithm}[!ht]
\caption{Sampling process for kernel hyperparameter $\phi$}
\label{alg:HyperSample}
\KwIn{$\phi$, $\boldsymbol{V},\boldsymbol{W},\boldsymbol{y}_{\Omega},\boldsymbol{\mathcal{X}},\tau$.}
\KwOut{next $\phi$.}
Initialize slice sampling scale $\rho=\log(10)$. \\
Compute $\boldsymbol{K}_{s}, \boldsymbol{K}_{\left.\boldsymbol{y}\right|\phi}, \log{p\left(\phi\given\boldsymbol{y}_{\Omega},\boldsymbol{V},\boldsymbol{W},\tau,\boldsymbol{\mathcal{X}}\right)}$ corresponding to $\phi$; \\
Calculate the sampling range: $\delta\sim\text{Uniform}(0,\rho)$, $\phi_{\min}=\phi-\delta$, $\phi_{\max}=\phi_{\min}+\rho$; \\
Draw $\eta\sim\text{Uniform}(0,1)$; \\
\While{True}{
Draw proposal $\phi'\sim\text{Uniform}(\phi_{\min},\phi_{\max})$; \\
Compute $\boldsymbol{K}_{s}', \boldsymbol{K}_{\left.\boldsymbol{y}\right|\phi'}, \log{p\left(\phi'\given\boldsymbol{y}_{\Omega},\boldsymbol{V},\boldsymbol{W},\tau,\boldsymbol{\mathcal{X}}\right)}$ corresponding to $\phi'$; \\
\uIf{$\exp{\big(\log{p\left(\phi'\given\boldsymbol{y}_{\Omega},\boldsymbol{V},\boldsymbol{W},\tau,\boldsymbol{\mathcal{X}}\right)}-\log{p\left(\phi\given\boldsymbol{y}_{\Omega},\boldsymbol{V},\boldsymbol{W},\tau,\boldsymbol{\mathcal{X}}\right)}\big)}>\eta$}
{\Return $\phi'$; \textbf{break};}
\uElseIf{$\phi'<\phi$}
    {$\phi_{\min}=\phi'$;}
\Else{$\phi_{\max}=\phi'$.}
}
\end{algorithm}

\subsection{Evaluation metrics}\label{appC}
The metrics for assessing the estimation accuracy on coefficients $\boldsymbol{\mathcal{B}}$ ($\text{MAE}_{\boldsymbol{\mathcal{B}}}/\text{RMSE}_{\boldsymbol{\mathcal{B}}}$) and unobserved output/response entries $\boldsymbol{y}_{\Omega^c}$ ($\text{MAE}_{\boldsymbol{y}_{\Omega^c}}/\text{RMSE}_{\boldsymbol{y}_{\Omega^c}}/\text{MAPE}_{\boldsymbol{y}_{\Omega^c}}$) are defined as:
\begin{equation*}
\begin{aligned}
&\text{MAE}_{\boldsymbol{\mathcal{B}}}=\frac{1}{MNP}\sum_{m=1}^{M}\sum_{n=1}^{N}\sum_{p=1}^{P}\left|b_{mnp}-\hat{b}_{mnp}\right|, \text{RMSE}_{\boldsymbol{\mathcal{B}}}=\sqrt{\frac{1}{MNP}\sum_{m=1}^{M}\sum_{n=1}^{N}\sum_{p=1}^{P}\left(b_{mnp}-\hat{b}_{mnp}\right)^2}, \\
&\text{MAE}_{\boldsymbol{y}_{\Omega^c}}=\frac{1}{MN-|\Omega|}\sum_{i\notin\Omega}\left|y_{i}-\hat{y}_{i}\right|, \text{RMSE}_{\boldsymbol{y}_{\Omega^c}}=\sqrt{\frac{1}{MN-|\Omega|}\sum_{i\notin\Omega}\left(y_{i}-\hat{y}_{i}\right)^2}, \\
&R^2=1-\frac{\sum_{i}\left(y_i-\hat{y}_i\right)^2}{\sum_{i}\left(y_i-\bar{y}\right)^2},
\end{aligned}
\end{equation*}
where $\hat{b}_{mnp}$ and $b_{mnp}$ are the $(m,n,p)$th element of the estimated and the setting/true coefficient tensor, respectively, $y_{i}$ and $\hat{y}_{i}$ are the actual value and estimation of $i$th unobserved data, respectively, and $\bar{y}$ denotes the average mean of the unobserved data. The criteria related to quantifying the uncertainty performance, i.e., CVG (interval coverage), INT (interval score), and CRPS (continuous rank probability score) of the 95\% posterior CI for the estimated $\boldsymbol{\mathcal{B}}$ values, are defined as follows:
\begin{equation} \notag
\begin{aligned}
\text{CVG}=&\frac{1}{MNP}\sum_{m=1}^{M}\sum_{n=1}^{N}\sum_{p=1}^{P}\mathbbm{1}\left\{{b}_{mnp}\in\left[l_{mnp},u_{mnp}\right]\right\}, \\
\text{INT}=&\frac{1}{MNP}\sum_{m=1}^{M}\sum_{n=1}^{N}\sum_{p=1}^{P}\left(u_{mnp}-l_{mnp}\right)+\frac{2}{\alpha}\left(l_{mnp}-b_{mnp}\right)\mathbbm{1}\{b_{mnp}<l_{mnp}\} \\
&~~~~~~~~~~~~~~~~~~~~~~~~~~~~~~~~~~~~~~~~~~~+\frac{2}{\alpha}\left(b_{mnp}-u_{mnp}\right)\mathbbm{1}\{b_{mnp}>u_{mnp}\}, \\
\text{CRPS}=&-\frac{1}{MNP}\sum_{m=1}^{M}\sum_{n=1}^{N}\sum_{p=1}^{P}\sigma_{mnp}\Bigg[\frac{1}{\sqrt{\pi}}-2\psi\left(\frac{b_{mnp}-\hat{b}_{mnp}}{\sigma_{mnp}}\right) \\
&~~~~~~~~~~~~~~~~~~~~~~~~~~~~~~~~~~~~~-\frac{b_{mnp}-\hat{b}_{mnp}}{\sigma_{mnp}}\left(2\Phi\left(\frac{b_{mnp}-\hat{b}_{mnp}}{\sigma_{mnp}}\right)-1\right)\Bigg], \\
\end{aligned}
\end{equation}
where $\psi$ and $\Phi$ denote the pdf (probability density function) and cdf (cumulative distribution function) of a standard normal distribution, respectively; $\sigma_{mnp}$ is the standard deviation (std.) of the estimation values after burn-in for $(m,n,p)$th entry of $\boldsymbol{\mathcal{B}}$, i.e., the std. of $\left\{\Tilde{\boldsymbol{\mathcal{B}}}^{(k)}\right\}_{k=K_1+1}^{K_1+K_2}$, $\alpha=0.05$, $[l_{mnp},u_{mnp}]$ denotes the 95\% central estimation interval for each coefficient value, and $\mathbbm{1}\{\cdot\}$ represents an indicator function that equals 1 if the condition is true and 0 otherwise.
\end{supplement}

\bibliographystyle{ba}
\bibliography{ref}

\end{document}